# ProbFlow: Joint Optical Flow and Uncertainty Estimation

Anne S. Wannenwetsch[1] Margret Keuper[2] Stefan Roth[1]
[1]TU Darmstadt [2]University of Mannheim

## Abstract

*Optical flow estimation remains challenging due to untextured areas, motion boundaries, occlusions, and more. Thus, the estimated flow is not equally reliable across the image. To that end, post-hoc confidence measures have been introduced to assess the per-pixel reliability of the flow. We overcome the artificial separation of optical flow and confidence estimation by introducing a method that jointly predicts optical flow and its underlying uncertainty. Starting from common energy-based formulations, we rely on the corresponding posterior distribution of the flow given the images. We derive a variational inference scheme based on mean field, which incorporates best practices from energy minimization. An uncertainty measure is obtained along the flow at every pixel as the (marginal) entropy of the variational distribution. We demonstrate the flexibility of our probabilistic approach by applying it to two different energies and on two benchmarks. We not only obtain flow results that are competitive with the underlying energy minimization approach, but also a reliable uncertainty measure that significantly outperforms existing post-hoc approaches.*

## 1. Introduction

Optical flow estimation has been extensively studied for more than three decades [5, 9, 14, 21, 36]. However, motion boundaries, large displacements, and occlusions still lead to erroneous flow fields, especially in challenging imaging conditions. In contrast, flow predictions are almost errorless in textured regions of uniform motion, which causes the reliability of optical flow predictions to vary greatly across the image. Uncertainty measures[1] aim to predict the reliability of the flow estimates and rate each estimate according to its predicted accuracy [24, 27, 30]. One application of such measures is to improve optical flow estimation using different ways of post-processing [4, 36]. Moreover, uncertainty measures represent an important tool when optical flow is used as a cue for other computer vision tasks, such as image segmentation or tracking, which consequently depend on the correctness of the flow. In such cases, error propagation can be avoided if only flow estimates of high confidence are taken into consideration, *e.g.* [33, 47].

[1] Uncertainty measures are closely related to confidence measures as their values are inversely related to confidence estimates.

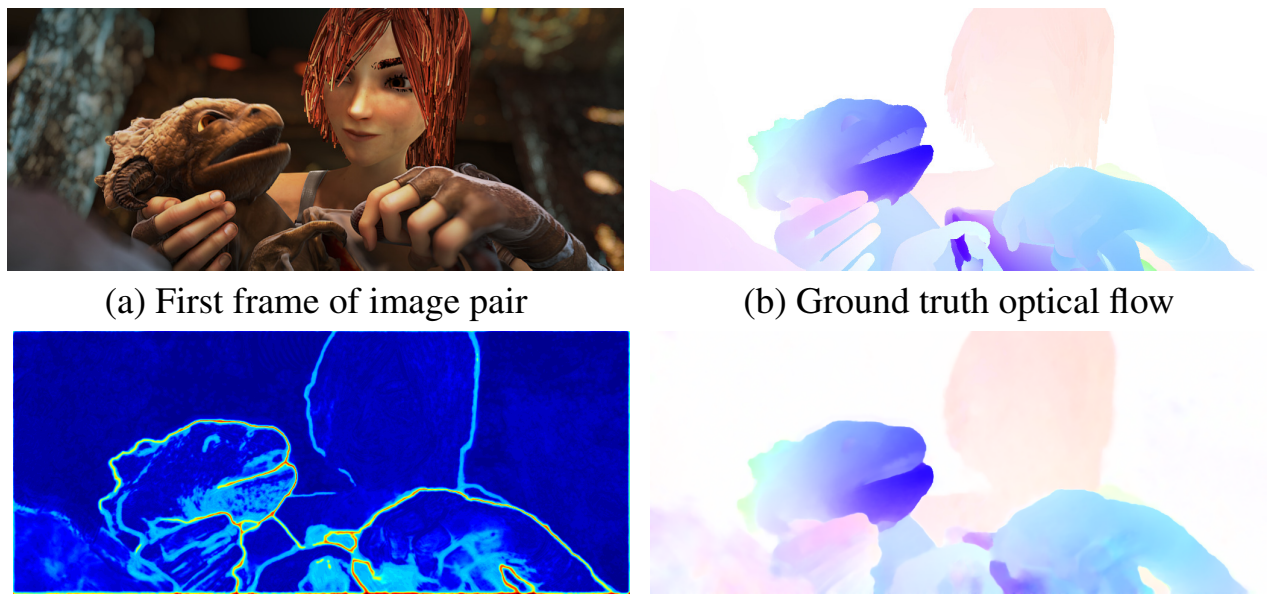

(a) First frame of image pair (b) Ground truth optical flow

(c) Uncertainty map from our method (d) Flow estimate from our method

Figure 1. Our uncertainty measure accurately detects regions with reliable flow predictions (dark blue) as well as parts of the image with erroneous estimates (dark red). *Best viewed on screen.*

Several post-hoc confidence measures have been introduced to directly estimate the uncertainty of flow predictions, *e.g.* [3, 20, 24, 30]. However, optical flow estimation is known to be rather challenging as a one-shot process [14, 15, 43]. We thus believe it is natural to ask why confidence estimation should be restricted to non-iterative (post-hoc) approaches. Moreover, flow algorithms for a concrete problem are carefully selected for the task at hand and provide us with an underlying model. We, thus, argue that it is very desirable to apply a *model-inherent* uncertainty measure tailored to the chosen method. We even go one step further and aim to *jointly predict* optical flow and a corresponding uncertainty in order to preserve and extract important information contained in the flow estimation process.

Our work focuses on energy-based methods [3, 15, 43] as they represent a flexible framework in widespread use. Starting from a general energy formulation, we derive a probability distribution of the flow field conditioned on the input images. In this setting, optical flow can be determined by minimizing the expected loss under the modeled posterior and the corresponding uncertainty measure is naturally obtained as the marginal entropy of the posterior at every pixel. Due to the intractability of exact posterior inference in general, we rely on variational approximate inference and use a mean-field approach. In this way, flow predictions and uncertainty estimations can be obtained as the result of a joint minimization problem that makes an additional training step of the uncertainty measure unnecessary [17, 30]. Fig. 1 shows an exemplary flow field and uncertainty prediction of our probabilistic approach obtained on the Sintel benchmark [11]. We term our approach *ProbFlow*.

To the best of our knowledge, we introduce the first *fully* probabilistic flow estimation approach that shows *competitive results* on established benchmarks. Moreover, our work is unique in that it is *broadly applicable* to a large number of existing energy-based optical flow algorithms and that we demonstrate how to benefit from the probabilistic framework in the context of uncertainty measures.

We apply ProbFlow to a classic Horn-Schunck-style energy as proposed in [43] and to the more recent EpicFlow formulation [36]. We show that the obtained flow fields are competitive with or even outperform the corresponding energy minimization results on the Middlebury [2] and Sintel [11] datasets. To assess the performance of the uncertainties estimated by our method, we rely on existing evaluation approaches and propose a new criterion based on the Spearman's rank correlation coefficient. Our uncertainty measure is clearly superior in comparison to various competing approaches and significantly outperforms the best existing methods. We further show the benefits of our uncertainty estimates in an application to motion segmentation [22] by generating highly reliable point trajectories.

## 2. Related Work

Starting from the work of Horn and Schunck [21], global energy minimization approaches have been used extensively for optical flow estimation; see [3, 15, 43] for in-depth overviews of existing methods and best practices. Despite their conceptual simplicity, global energy formulations are still up-to-date [13, 29]. Moreover, energy-based approaches are frequently used for post-processing optical flow predictions obtained through other means [1, 14, 36].

In the past, a variety of optical flow uncertainty measures have been proposed. Various simple methods are based on the characteristics of the input data only, *e.g.* [3, 20]. We refer to [24, 27] for a more comprehensive summary of such confidence approaches. In comparison to our work, all of these measures omit important sources of information by not considering the estimated optical flow field itself.

A second class of confidence measures relies on an analysis of the estimated flow fields. Kondermann *et al.* [23] learn a linear subspace of true spatio-temporal flow neighborhoods and use the reconstruction error of an estimated flow vector to evaluate its reliability. In [24], a probabilistic model of flow patches is learned and approximated with a Gaussian distribution, which yields a confidence measure based on hypothesis testing. But this second class of confidence measures does not consider all aspects of flow uncertainty either, as the input images are not taken into account.

Recently, Mac Aodha *et al.* [30] proposed a confidence measure based on the input as well as the output of an optical flow algorithm. Using a multi-cue feature vector, a classifier is trained to predict whether the endpoint error at a certain pixel is smaller than a previously defined threshold. The probability output of the classifier is used as a confidence measure. However, [30] does not take advantage of uncertainty information available in the flow estimation procedure itself and requires a separate training step that is unnecessary for our model-inherent approach.

Uncertainty measures tailored to energy minimization approaches have been proposed especially for *local* energy methods, *e.g.*, [3, 31, 40, 41]. Model-inherent confidence measures designed for *global* energy formulations are quite rare, on the other hand. Kybic and Nieuwenhuis [27] introduce a method that relies on bootstrap resampling. Based on varying pixel contributions, repeated optical flow estimation is performed and a confidence measure is determined as the total standard deviation of the obtained flow predictions. Bruhn and Weickert [8] propose the inverse local energy as a confidence measure, which is applicable to a broad variety of energies. A large uncertainty is thus associated with a strong violation of the local model assumptions encoded in the energy formulation. Gehrig and Scharwächter [17] use the energy and combine it with several features such as the spatial and temporal flow variance in order to obtain a real-time confidence estimate. In contrast to the two above approaches, our method does not explicitly limit the receptive field of the confidence measure by only considering the local neighborhood of a pixel. Instead, uncertainty propagation is facilitated by means of iterative spatial inference.

In the past, several methods based on a probabilistic flow formulation have been proposed [35, 39, 40, 41]. However, many of these works are based on simplifying assumptions such as locally constant flow or a Gaussian distribution of brightness constancy errors. In [19, 26, 44], probabilistic approaches are applied to obtain improved models of optical flow. In contrast to our work, these methods do not apply a fully probabilistic approach, but fall back to a maximum a-posteriori (MAP) estimate, *i.e.* minimize the underlying energy. Glocker *et al.* [18] use flow uncertainties in a dynamic Markov random field but rely on a discrete approach. The most closely related work of Chantas *et al.* [12] applies a variational-Bayes approach similar to ours. However, their method is not designed as a stand-alone flow algorithm, but only as an improved initialization in comparison to a simple Horn-Schunck approach. Therefore, the obtained results are not competitive with respect to the state of the art. Moreover, the paper does not take advantage of the probabilistic approach in the context of uncertainty measures.

## 3. Energy Framework

Since our joint, model-inherent uncertainty estimation developed below is broadly applicable to different kinds of energy-based optical flow approaches [3, 15, 43], we first introduce a formalization that allows us to describe previously proposed optical flow methods in a unified manner.

In the following, we estimate optical flow between an image pair $I = \{I_1, I_2\}$ and denote the estimate as $\mathbf{y} = (\mathbf{y}_{ij})_{ij} = \left(u_{ij}, v_{ij}\right)^{\mathrm{T}}_{ij}$ for pixels $(i,j)$, $i = 1,\dots,n$, $j = 1,\dots,m$. Energy-based approaches estimate the optical flow as the minimizer $\mathbf{y}^\star$ of an energy function

$$E\left(\mathbf{y}; I\right) = E_{\mathrm{D}}\left(\mathbf{y}; I\right) + \lambda_{\mathrm{S}} E_{\mathrm{S}}\left(\mathbf{y}\right) \tag{1}$$

with $E_{\mathrm{D}}\left(\mathbf{y}; I\right)$ denoting a data term that encourages the flow to be consistent with input images $I_1$ and $I_2$. The spatial term $E_{\mathrm{S}}\left(\mathbf{y}\right)$ imposes a (smoothness) prior on the flow, and $\lambda_{\mathrm{S}}$ represents a trade-off parameter between the terms.

In the following, we use functions $f_{\mathrm{D}}(\cdot)$ and $f_{\mathrm{S}}(\cdot)$ to formalize violations of the assumptions underlying the chosen optical flow model. For instance, the intensity difference of corresponding pixels in $I_1$ and $I_2$ may be evaluated to model a brightness constancy assumption for the data term. So-called penalty functions $\rho_{\mathrm{D}}(\cdot)$ and $\rho_{\mathrm{S}}(\cdot)$ penalize violations of the assumptions modeled in $f_{\mathrm{D}}(\cdot)$ and $f_{\mathrm{S}}(\cdot)$. The energy terms $E_{\mathrm{D}}$ and $E_{\mathrm{S}}$ can then be described as a sum of the contributions from all pixels such that

$$E_{\mathrm{D}}\left(\mathbf{y}; I\right) = \sum_{i,j} \rho_{\mathrm{D}}\Big(f_{\mathrm{D}}\left(\mathbf{y}_{ij}; I\right)\Big) \tag{2}$$

$$E_{\mathrm{S}}\left(\mathbf{y}\right) = \sum_{i,j} \sum_{(i',j') \in S(i,j)} \rho_{\mathrm{S}}\Big(f_{\mathrm{S}}\left(\mathbf{y}_{ij}, \mathbf{y}_{i'j'}\right)\Big) \tag{3}$$

with $S(i,j)$ describing a set of neighbors of pixel $(i,j)$.

It has been shown that the estimated flow field $\mathbf{y}^\star$ can be improved by adding a non-local term to the energy function in Eq. (1) [25, 43]. Thus, we optionally use an additional non-local term $E_{\mathrm{N}}\left(\mathbf{y}\right)$ akin to $E_{\mathrm{S}}\left(\mathbf{y}\right)$, considering an extended neighborhood $N(i,j)$. Again, $E_{\mathrm{N}}\left(\mathbf{y}\right)$ is described by a model assumption $f_{\mathrm{N}}(\cdot)$ and its corresponding penalty function $\rho_{\mathrm{N}}(\cdot)$.

Common choices for penalty functions $\rho(\cdot)$ are, *e.g.*, a quadratic function [21], a Lorentzian function [5], or a (generalized) Charbonnier function [9, 43]. Many of these functions can be described by the negative logarithm of a Gaussian Scale Mixture (GSM) [44, 46]. Thus, in the following we rely on this simple but powerful class of functions and represent the penalty terms as GSMs of $L$ components

$$\rho\left(z\right) = -\log\left[\sum_{l=1}^{L} \pi_l \, \mathcal{N}\left(z; 0, \sigma_l^2\right)\right] \tag{4}$$

with $\mathcal{N}\left(z; \mu, \sigma\right)$ being a normal distribution with mean $\mu = 0$ and variance $\sigma = \sigma_l$; the weights $\pi_l \geq 0$ sum to 1. As we will see, GSMs also benefit our probabilistic approach.

## 4. Probabilistic Interpretation and Inference

We now aim to approach optical flow estimation in a probabilistic manner. As the energy function $E\left(\mathbf{y}; I\right)$ from Eq. (1) describes a Markov random field, it is easy to derive the corresponding posterior distribution in its Gibbs form as

$$p\left(\mathbf{y} \mid I\right) = \frac{1}{Z} \exp\left\{-\frac{1}{T} E(\mathbf{y}; I)\right\} \tag{5}$$

with partition function $Z \equiv Z(I, T, \lambda_{\mathrm{S}}, \lambda_{\mathrm{N}})$ and temperature $T$. In the following, w.l.o.g. we set $T = 1$ and introduce an additional parameter $\lambda_{\mathrm{D}}$ scaling the data term $E_{\mathrm{D}}$.

To ease probabilistic inference, we use the same procedure as [43] and introduce an auxiliary flow field $\hat{\mathbf{y}}$ that allows us to decouple the non-local potential from the remaining terms of the posterior distribution in Eq. (5), *i.e.*

$$\begin{aligned} p\left(\mathbf{y}, \hat{\mathbf{y}} \mid I\right) = \frac{1}{Z} \exp\Big\{ & -\lambda_{\mathrm{D}} E_{\mathrm{D}}(\mathbf{y}; I) - \lambda_{\mathrm{S}} E_{\mathrm{S}}(\mathbf{y}) \\ & - \lambda_{\mathrm{C}} E_{\mathrm{C}}\left(\mathbf{y}, \hat{\mathbf{y}}\right) - \lambda_{\mathrm{N}} E_{\mathrm{N}}\left(\hat{\mathbf{y}}\right) \Big\} \end{aligned} \tag{6}$$

with $E_{\mathrm{C}}\left(\mathbf{y}, \hat{\mathbf{y}}\right) = \sum_{i,j} \|\mathbf{y}_{i,j} - \hat{\mathbf{y}}_{i,j}\|_2^2$.

The log-posterior of Eq. (6) now has terms in the form of the logarithm of a sum of exponentials, which is challenging when deriving closed-form mean-field updates below. Here, we benefit from our choice to model each penalty function $\rho(\cdot)$ as a GSM. We follow [16, 28] to retain explicit latent variables $\mathbf{h} = (\mathbf{h}_{\mathrm{D}}, \mathbf{h}_{\mathrm{S}}, \mathbf{h}_{\mathrm{N}})$, which are chosen to follow a discrete distribution and to have a 1-of-L representation. We then obtain an augmented penalty function $\rho_\gamma\left(z, \mathbf{h}_\gamma\right)$ with $\gamma \in \{\mathrm{D}, \mathrm{S}, \mathrm{N}\}$ as

$$\rho_\gamma\left(z, \mathbf{h}_\gamma\right) = -\log\left[\prod_{l=1}^{L} \pi_l^{h_{\gamma,l}} \, \mathcal{N}\left(z; 0, \sigma_l^2\right)^{h_{\gamma,l}}\right]. \tag{7}$$

At this point, the posterior $p(\mathbf{y}, \hat{\mathbf{y}}, \mathbf{h} \mid I)$ represents the probabilistic equivalent of the energy $E\left(\mathbf{y}; I\right)$ in Eq. (1).

In our probabilistic setup, the flow estimate $\mathbf{y}^\star$ is chosen to minimize the expected loss over $p(\mathbf{y}, \hat{\mathbf{y}}, \mathbf{h} \mid I)$, *i.e.*

$$\mathbf{y}^\star = \arg\min_{\tilde{\mathbf{y}}} \mathbb{E}_{p(\mathbf{y}, \hat{\mathbf{y}}, \mathbf{h} \mid I)}\left[l(\mathbf{y}, \tilde{\mathbf{y}})\right] \tag{8}$$

with $l\left(\cdot, \cdot\right)$ being a suitable loss function such as the Average End-Point Error (AEPE) [34]. The desired uncertainty measure can be obtained by considering the marginal distribution $p(\mathbf{y}_{ij} \mid I)$ of the flow estimates. In particular, we propose to use the marginal entropy at every pixel as a model-inherent uncertainty estimate of the flow prediction.

**Inference.** To obtain a flow estimate and the underlying marginal distributions at every pixel, we rely on an approximate inference scheme. We use variational inference [45][2] and approximate $p(\mathbf{y}, \hat{\mathbf{y}}, \mathbf{h} \mid I)$ with the help of a tractable distribution $q(\mathbf{y}, \hat{\mathbf{y}}, \mathbf{h}; \boldsymbol{\theta})$ and variational parameters $\boldsymbol{\theta}$.

[2]Not to be confused with variational formulations common in flow.

Following a naive mean-field assumption, we choose the parametric distribution $q$ to be factorized over $\mathbf{y}$, $\hat{\mathbf{y}}$, as well as $\mathbf{h}$. The marginal distribution of flow vectors $\mathbf{y}_{ij}$ and $\hat{\mathbf{y}}_{ij}$ is assumed to be Gaussian, *e.g.*

$$q\left(\mathbf{y}_{ij}; \boldsymbol{\theta}\right) \sim \mathcal{N}\left(\mathbf{y}_{ij}; \boldsymbol{\mu}_{ij}, \boldsymbol{\Sigma}_{ij}\right). \tag{9}$$

It is reasonable to assume the horizontal and vertical flow components to be uncorrelated [37]. Thus, the covariances $\boldsymbol{\Sigma}_{ij}$ are modeled as diagonal matrices. As in [16, 28], the distributions of the latent variables are chosen to be multinomial

$$q(\mathbf{h}_{\gamma,ij}; \boldsymbol{\theta}) = \prod_{l=1}^{L} k_{\gamma,ij,l}^{h_{\gamma,ij,l}} \tag{10}$$

so that the parameters $k_{\gamma,ij,l} \geq 0$ satisfy $\sum_l k_{\gamma,ij,l} = 1$. The approximating distribution $q$ is then given as

$$q(\mathbf{y}, \hat{\mathbf{y}}, \mathbf{h}; \boldsymbol{\theta}) = \prod_{i,j} \Big[ q\left(\mathbf{y}_{ij}; \boldsymbol{\theta}\right) \cdot q\left(\hat{\mathbf{y}}_{ij}; \boldsymbol{\theta}\right) \cdot \prod_{\gamma \in \{\mathrm{D,S,N}\}} q\left(\mathbf{h}_{\gamma,ij}; \boldsymbol{\theta}\right) \Big] \tag{11}$$

with $\boldsymbol{\theta} = \left\{ \boldsymbol{\mu}_{ij}, \hat{\boldsymbol{\mu}}_{ij}, \boldsymbol{\Sigma}_{ij}, \hat{\boldsymbol{\Sigma}}_{ij}, \mathbf{k}_{\gamma,ij} \right\}_{ij,\gamma}$.

Suitable variational parameters $\boldsymbol{\theta}^\star$ of $q$ are determined such that the Kullback-Leibler (KL) divergence between $p$ and its approximating distribution $q$ is minimized, *i.e.*

$$\boldsymbol{\theta}^\star = \arg\min_{\boldsymbol{\theta}} D_{KL}\big(q(\mathbf{y}, \hat{\mathbf{y}}, \mathbf{h}; \boldsymbol{\theta}) \mid p(\mathbf{y}, \hat{\mathbf{y}}, \mathbf{h} \mid I)\big) \tag{12}$$

$$= \arg\min_{\boldsymbol{\theta}} \mathbb{E}_{q(\mathbf{y},\hat{\mathbf{y}},\mathbf{h};\boldsymbol{\theta})} \big[ \log q(\mathbf{y}, \hat{\mathbf{y}}, \mathbf{h}; \boldsymbol{\theta}) \big] - \mathbb{E}_{q(\mathbf{y},\hat{\mathbf{y}},\mathbf{h};\boldsymbol{\theta})} \big[ \log p(\mathbf{y}, \hat{\mathbf{y}}, \mathbf{h} \mid I) \big]. \tag{13}$$

Due to the usage of explicit latent variables $\mathbf{h}$, it is possible to compute the expectations in Eq. (13) in an analytic way. While the derivation of the corresponding equations is tedious, the individual steps are elementary; see supplemental material for a more detailed explanation of the procedure.

We now estimate the flow by replacing the posterior $p$ in Eq. (8) with its approximating distribution $q$. When performing Bayesian risk minimization of the AEPE, the optical flow prediction is obtained as $\mathbf{y}^\star_{ij} = \boldsymbol{\mu}_{ij}$ and, therefore, corresponds to the mode of the variational distribution $q$.

As $q(\mathbf{y}, \hat{\mathbf{y}}, \mathbf{h}; \boldsymbol{\theta})$ was defined in a factorized way, the corresponding marginal distribution at pixel $(i, j)$ is given as $q\left(\mathbf{y}_{ij}; \boldsymbol{\theta}\right)$ and our proposed model-inherent uncertainty measure can be obtained as the marginal entropy

$$\Psi_{\text{ProbFlow}} = H(\mathbf{y}_{ij}) = \log\left(\det\left(\boldsymbol{\Sigma}_{ij}\right)\right) + \text{const}. \tag{14}$$

# 5. Specific Models

We now apply our ProbFlow approach to two specific energy functions commonly used for optical flow estimation.

## 5.1. Probabilistic Classic Flow

We first consider a classical Horn-Schunck-based objective based on brightness constancy as given in [43]. Following the common approach to use a first-order Taylor approximation for a linearization of the data term, we obtain

$$f_{\mathrm{D}}\left(\mathbf{y}_{ij}; I\right) = I_2\left(i + u^0_{ij}, j + v^0_{ij}\right) - I_1\left(i, j\right) + \nabla_2 I_2\left(i + u^0_{ij}, j + v^0_{ij}\right)^{\mathrm{T}} \left(\mathbf{y}_{ij} - \mathbf{y}^0_{ij}\right), \tag{15}$$

where $\mathbf{y}^0_{ij}$ is the point of approximation and $\nabla_2 I_2$ denotes the spatial derivatives of $I_2$. The smoothness prior in [43] assumes small flow gradients over a 4-neighborhood of horizontal and vertical flow components. The non-local term encourages smoothness over a neighborhood of size $5 \times 5$. In both cases, the smoothness assumption is represented as

$$f_{\mathrm{S}}\left(x_{ij}, x_{i'j'}\right) = f_{\mathrm{N}}\left(x_{ij}, x_{i'j'}\right) = x_{ij} - x_{i'j'}. \tag{16}$$

Akin to energy minimization approaches [43], we found the use of the non-local term crucial for obtaining accurate optical flow estimates, since mean-field inference using only the terms $E_{\mathrm{D}}$ and $E_{\mathrm{S}}$ is rather outlier-prone.

## 5.2. Probabilistic FlowFields

In a second setup, we aim for a probabilistic version of FlowFields [1]. We follow Bailer *et al.* and consider the energy used in the post-processing step of EpicFlow [36], which uses a data term based on gradient constancy. As proposed in [49], the image gradient terms are normalized w.r.t. the spatial derivatives, which is helpful to avoid outliers in the flow field. We obtain the linearized data assumption as

$$f_{\mathrm{D}}\left(\mathbf{y}_{ij}; I\right) = \left\| \sum_{r=1}^{3} \theta^r_{ij} \circ \Big[ \nabla_2 I^r_2\left(i + u^0_{ij}, j + v^0_{ij}\right) - \nabla_2 I^r_1\left(i, j\right) + \mathbf{H}\Big(I^r_2\left(i + u^0_{ij}, j + v^0_{ij}\right)\Big)\left(\mathbf{y}_{ij} - \mathbf{y}^0_{ij}\right) \Big] \right\|_2. \tag{17}$$

Here, $r = 1, \ldots, 3$ indicates the RGB color channels, $\mathbf{H}\left(I^r_2\right)$ denotes the Hessian of $I^r_2$, and $\circ$ is the Hadamard product. The normalization coefficient $\theta^r_{ij}$ is given as

$$\theta^r_{ij} = \begin{pmatrix} \theta^r_{ij,1} \\ \theta^r_{ij,2} \end{pmatrix}, \; \theta^r_{ij,k} = \frac{1}{\sqrt{\|\mathbf{H}(I^r_2) \cdot \mathbf{e}_k\|^2_2 + \zeta^2}} \tag{18}$$

with 2D unit vectors $\mathbf{e}_k$ and a small constant $\zeta > 0$ [49].

The smoothness term of [36] is based on the flow gradient norm and uses additional filters of size $2 \times 3$ and $3 \times 2$ as described in [38] when calculating the flow derivatives. To keep the inference problem consistent, our approach differs slightly from [36] in that we only use the forward gradient filter described in Eq. (16) and obtain

$$f_{\mathrm{S}}\left(\mathbf{y}_{ij}, \mathbf{y}_{i'j'}\right) = \sqrt{\left(u_{ij} - u_{i'j'}\right)^2 + \left(v_{ij} - v_{i'j'}\right)^2} \tag{19}$$

for $\mathbf{y}_{i'j'}$ in a 4-neighborhood of $\mathbf{y}_{ij}$.

Following [36], a locally adaptive trade-off parameter $\lambda_{\text{S}}(\mathbf{x}) = \exp(-\kappa \|\nabla_2 I_1(\mathbf{x})\|)$ is used. Similar to MAP estimation, the minimization of the KL divergence in Eq. (13) obtained from $f_{\text{D}}$, $f_{\text{S}}$, and $\lambda_{\text{S}}(\mathbf{x})$ performs well in practice. Therefore, an additional non-local term can be neglected.

## 6. Implementation

Optical flow estimation by energy minimization is known to be far from easy. Similarly, an application of mean-field inference is non-trivial in this context. Moreover, it is essential to consider several details commonly used with energy-based approaches to obtain satisfying results [15, 43]. A summary of basic design choices as well as an extensive analysis of the influence of all described specifics can be found in the supplemental material.

Common per-pixel updates, *e.g.* [48], do not work well for the mean-field inference of Eq. (13). Instead, we keep the corresponding optimization procedure as efficient as possible and use a block-coordinate descent scheme updating flow estimates $\boldsymbol{\mu}$, variances $\boldsymbol{\Sigma}$, and latent variables $\mathbf{k}$ in an alternating manner. We derive the gradient of the KL divergence in Eq. (13), set it to zero, and obtain an update equation for each set of variables (see supplemental). As with MAP, we found a joint update of the flow predictions at all pixels to be crucial to obtaining smooth flow fields.

**Parameters.** To determine suitable trade-off parameters $\lambda$, it is not sufficient to follow the common approach and choose parameters that lead to the smallest AEPE on a training set [15, 43]. Instead, we evaluate the quality of both the obtained flow predictions and the uncertainty measure in order to ensure accurate flow estimates *and* meaningful entropies. To assess the quality of our uncertainty measure, we follow the approaches in [8, 24, 27, 30] and compute so-called sparsification plots. To that end, the pixels of a flow field are sorted according to the estimated uncertainties. Subsequently, an increasing percentage of the pixels is removed and the AEPE of the remaining pixels is calculated. In order to evaluate how well different uncertainty measures perform on an entire dataset, we propose to normalize the graphs, calculate the area under curve (AUC), and average over the sequences. To consider the trade-off between flow accuracy and quality of the uncertainty measure, we evaluate an $\text{F}_1$-score

$$\text{F}_1 = \frac{\text{AEPE} \cdot c \ \text{AUC}}{\text{AEPE} + c \ \text{AUC}} \tag{20}$$

over the training data with constant $c$ weighting the influence of the two metrics. We then determine parameters using Bayesian optimization [42] of Eq. (20).

Concerning the penalty functions, we follow the approach in [44] and learn appropriate GSM models for data likelihood, smoothness prior, as well as the non-local term from the respective training datasets or a randomly chosen subset thereof. Using manually determined variances $\sigma_l$, a simple expectation maximization algorithm is used to obtain the corresponding weights $\pi_l$. We observe that GSMs with $L = 10$ components perform well in practice. To save computational time, we resort to $L = 5$ components for our probabilistic implementation of Classic Flow.

**Details.** For Probabilistic Classic Flow, we follow the underlying energy approach (ClassicA, [43]) and use zero flow as an initialization; we denote the method as *ProbClassicA*. Variances are initialized as $\sigma_{\text{init}} = 1\text{e}{-7}$, latent variables as $k_{\text{init}} = 1/L$. During the inference process, the parameter $\lambda_{\text{C}}$ is annealed as in [43]. The parameters $\lambda_{\text{D}}$, $\lambda_{\text{S}}$, and $\lambda_{\text{N}}$ are obtained with Bayesian optimization of the $\text{F}_1$-score (Eq. 20) using $c = 5$. As suggested by Sun *et al.* in the context of MAP, we use the variables $\hat{\boldsymbol{\mu}}$ as the flow prediction. The corresponding uncertainties are then obtained from $\hat{\boldsymbol{\Sigma}}$.

For Probabilistic FlowFields, we use the state-of-the-art FlowFields method [1] to generate sparse matches. Following Bailer *et al.*, we apply the EpicFlow [36] post-processing step with Sintel parameters to interpolate the matches and use the results to initialize our algorithm. This method will be denoted as *ProbFlowFields*. Variances and latent variables are initialized as for ProbClassicA. Trade-off parameters $\lambda_{\text{D}}$, $\lambda_{\text{S}}$, and $\kappa$ are obtained again with Bayesian optimization [42]. The parameter of the $\text{F}_1$-score (Eq. 20) is chosen as $c = 100$ since the AEPE is significantly higher on Sintel. We will make code available for ProbClassicA as well as for ProbFlowFields.

## 7. Experiments & Results

In the following, we evaluate our probabilistic flow approach and assess the quality of our uncertainty measure by comparing it to different approaches from the literature.

### 7.1. Competing uncertainty measures

We apply existing uncertainty measures on top of the corresponding energy minimization approaches (*i.e.* ClassicA or FlowFields) in order to analyze the benefit of our combined flow prediction and uncertainty estimation. We limit ourselves to a few select methods here and give a more extensive comparison in the supplemental material.

Barron *et al.* [3] suggest a confidence measure based on the spatial gradient of the input image. The corresponding uncertainty is obtained as $\Psi_{\text{Gradient}} = -\|\nabla_2 I_1\|$ using central differences to approximate the gradient.

The learned confidence measure in [30] is based on a classifier that predicts with probability $\tilde{p}$ whether the error of the flow estimate at a certain pixel is smaller than a specified threshold $\epsilon_T$. We use the implementation of MacAodha *et al.* and train $\Psi_{\text{Learned}} = -\tilde{p}$ as described in [30].

| | training | | test | |
|---|---|---|---|---|
| Method | AEPE | rel. chg. | AEPE | rel. chg. |
| Classic++ [43] | **0.285** | **-0.04** | **0.406** | **-0.07** |
| ClassicA [43] | 0.295 | >-0.01 | – | – |
| ProbClassicA *(ours)* | 0.296 | 0.00 | 0.435 | 0.00 |

Table 1. Average end-point error (AEPE) and its relative change (rel. chg.) in comparison to ProbClassicA on Middlebury.

As proposed by Bruhn and Weickert [8], we use the local energy contribution of the MAP estimate $\mathbf{y}_{\text{MAP}}$ as a model-inherent uncertainty $\Psi_{\text{Energy}} = E(\mathbf{y}_{\text{MAP}}; I)$. We apply the non-linear version of the underlying energy function as it shows an improved performance [10]. A data penalty for out-of-boundary pixels is determined on the training set.

For an additional baseline, we perform a Laplace approximation of the posterior $p(\mathbf{y}, \hat{\mathbf{y}} \mid I)$ around $\mathbf{y}_{\text{MAP}}$. With $\mathbf{H}$ denoting the Hessian of the linearized energy $E(\mathbf{y}_{\text{MAP}}; I)$, the covariance is given as $\mathbf{\Sigma}_L = \mathbf{H}^{-1}$. Similar to our approach, we obtain $\Psi_{\text{Laplace}} = -\log\big(\det(\mathbf{H})\big) + \text{const}$.

Finally, we consider an oracle uncertainty measure $\Psi_{\text{Oracle}}$, for which the uncertainty of a pixel is given by its end-point error. Thus, the estimate $\Psi_{\text{Oracle}}$ provides a bound for the best possible uncertainty estimation. Note that we consider oracle uncertainties from the predictions obtained by MAP estimation; the oracle uncertainties for our flow predictions show a very similar behavior.

### 7.2. ProbClassicA

We first evaluate the application of ProbFlow to the ClassicA model described in Sec. 5.1. We rely on the Middlebury dataset [2], which has frequently been used to evaluate the performance of optical flow confidence measures. In order to compare different uncertainty approaches, ground truth optical flow is needed. As the Middlebury training set includes only 8 image pairs, we need to resort to training and testing the uncertainty measures on the same data.

To compare the performance of different flow estimation algorithms, we rely on the commonly used AEPE. Table 1 gives results on Middlebury training and test. Our method (ProbClassicA) performs on par with ClassicA. On the one hand, this is to be expected as both approaches share the same energy formulation and energy-based methods use highly elaborate schemes. On the other hand, this is the first time that a fully probabilistic method achieves competitive results on a public benchmark. For completeness, we also include the related Classic++, which shows slightly better results as its median filtering step allows for a more effective outlier suppression than an additional nonlocal term [25]. The median filter cannot easily be applied to our case, but the results for ClassicA and ProbClassicA could be further improved by adding weights to the nonlocal term [43].

The evaluation of uncertainty measures is more challenging. Sparsification plots are commonly considered [8, 24, 27, 30]. To be able to compare the performance of different uncertainty measures over an entire dataset, we calculate the AUC as described in Sec. 6. However, as argued by Márquez-Valle *et al.* [31], sparsification plots do not allow to estimate how strongly an uncertainty estimate is related to the underlying pixel errors. To address this, we propose to compute the Spearman's rank correlation coefficient (CC), which estimates how well the examined uncertainty values can be mapped onto the corresponding end-point errors using an arbitrary monotonic function.

| Uncertainty measure | AUC | rel. chg. | CC | rel. chg. |
|---|---|---|---|---|
| Gradient [3] | 0.971 | 1.08 | 0.023 | 0.94 |
| Laplace | 0.656 | 0.41 | 0.160 | 0.57 |
| Energy [8] | 0.498 | 0.07 | 0.303 | 0.19 |
| Learned [30] | 0.496 | 0.06 | 0.324 | 0.13 |
| ProbClassicA *(ours)* | **0.466** | **0.00** | **0.374** | **0.00** |
| Oracle | *0.255* | – | *1.000* | – |

Table 2. Area under curve (AUC), Spearman's rank correlation coefficient (CC), and relative change (rel. chg.) in comparison to our uncertainty measure on the Middlebury dataset.

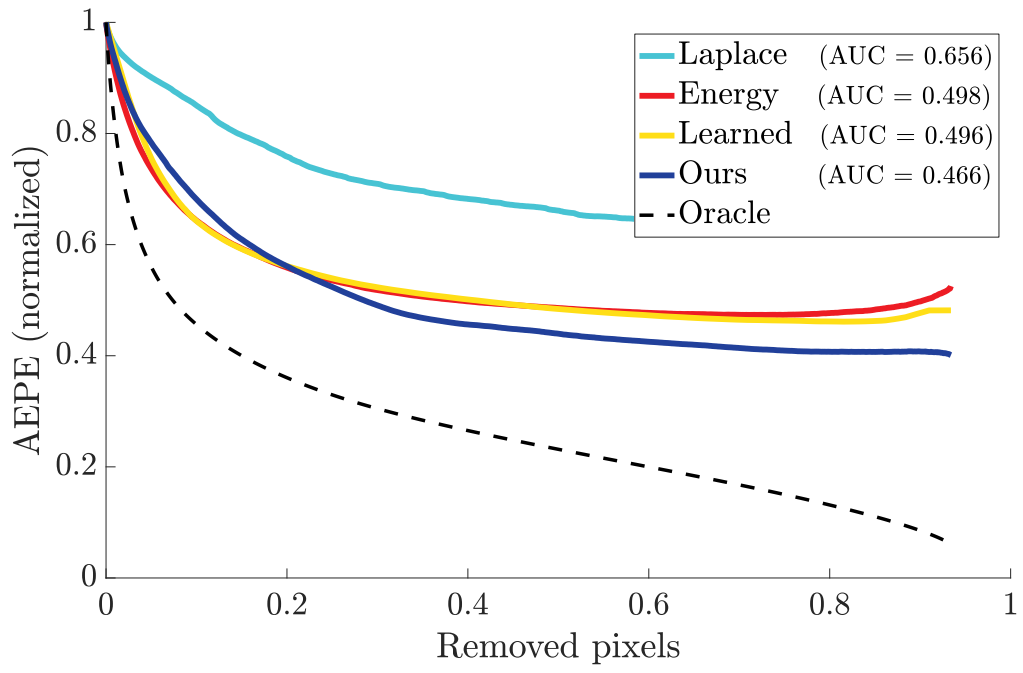

Figure 2. Sparsification plots averaged over Middlebury training.

Table 2 and the sparsification plots in Fig. 2 show that our uncertainty predictions are clearly superior in detecting the most reliable flow estimates. The gradient uncertainty has almost no ability to rank the pixels according to their accuracy. Hence, the simple consideration of input data appears insufficient. Similarly, the Laplace measure does not lead to satisfying results. $\Psi_{\text{Energy}}$ and $\Psi_{\text{Learned}}$ lead to very similar AUC results, whereby the learned uncertainty takes substantial time for training. Our method improves the AUC over previous ones by more than 6%. Moreover, the evaluation of the CC reveals that our uncertainty measure shows by far the highest correlation between the assigned uncertainty value and the per-pixel end-point error with a relative improvement of 13% over the learned uncertainty.

### 7.3. ProbFlowFields

Next, we apply our probabilistic approach to the competitive FlowFields method as described in Sec. 6. We use the more recent Sintel benchmark [11], which in comparison to the Middlebury dataset, allows to partition its more than 1000 flow sequences into training and validation sets. Exemplary flow and uncertainty estimates as well as the corresponding ground truth can be seen in Fig. 3.

| Method | validation | | test | |
|---|---|---|---|---|
| | AEPE | rel. chg. | AEPE | rel. chg. |
| Initialization | 3.303 | 0.06 | – | – |
| FlowFields [1] | 3.147 | <0.01 | 5.727$^{\dagger}$ | <0.01 |
| FlowFields$^{\star}$ | 3.161 | 0.01 | – | – |
| ProbFlowFields *(ours)* | 3.127 | 0.00 | 5.696 | 0.00 |
| ProbFlowFields + BS | **3.052** | **-0.02** | **5.628** | **-0.01** |

Table 3. Average end-point error (AEPE) and relative change (rel. chg.) w.r.t. to ProbFlowFields on Sintel. $^{\dagger}$FlowFields shows better results than the ones published on the website. See text for details.

| Uncertainty measure | AUC | rel. chg. | CC | rel. chg. |
|---|---|---|---|---|
| Gradient [3] | 1.022 | 1.57 | -0.009 | 1.02 |
| Laplace | 0.657 | 0.65 | 0.257 | 0.54 |
| Energy [8] | 0.470 | 0.18 | 0.434 | 0.23 |
| Learned [30] | 0.474 | 0.19 | 0.451 | 0.20 |
| ProbFlowFields *(ours)* | **0.398** | **0.00** | **0.563** | **0.00** |
| Oracle | *0.182* | – | *1.000* | – |

Table 4. Area under curve (AUC), Spearman's rank correlation coefficient (CC), and relative change (rel. chg.) in comparison to our uncertainty measure on a Sintel benchmark validation set.

Table 3 summarizes the AEPEs on our validation set and the test set of the Sintel benchmark. For comparison, we also report results of *FlowFields*$^{\star}$, based on the same consistent EpicFlow energy variant underlying ProbFlowFields, *c.f*. Sec. 5.2. Again, our estimates from ProbFlowFields are competitive with its underlying energy method and we observe results on par with the original FlowFields.

Evaluated on the Sintel test set, ProbFlowFields currently ranks 6$^{\text{th}}$ in comparison to previously published methods. Please note that the FlowFields test results shown in Table 3 are superior to the publicly available AEPE of 5.810. To suppress the effect of the random component in the matches of FlowFields and thus have a fair comparison, we have re-evaluated the original FlowFields implementation using the exact same matches as for our approach.

The performance of the competing uncertainty measures is evaluated in Table 4. Again, the Gradient and Laplace uncertainties perform considerably worse than the remaining approaches. The measures $\Psi_{\text{Learned}}$ and $\Psi_{\text{Energy}}$ result in similar AUC values. Our method again leads to a clear improvement of over 18%. Moreover, we also improve the CC by 20% in comparison to the second best measure. The sparsification plots in Fig. 4 show that our uncertainty measure leads to the best result for all fractions of removed pixels. Strikingly, the energy-based uncertainty as well as the Laplace measure show a strong increase of the AEPE when only a small fraction of pixels are kept. This is clearly undesirable as it indicates that the optical flow predictions are incorrect for pixels that are considered as highly reliable. Our probabilistic approach does not show this behavior.

To illustrate the benefits of our uncertainties in post-processing, we apply the fast bilateral solver [4] on top of ProbFlowFields, improving the AEPE by 2.4% and 1.2% on the validation and test set, respectively (*c.f*. Table 3). In comparison, post-processing assuming equally reliable flow yields an improvement of only 0.4% on the validation set.

In a last experiment, we evaluated the overall runtime on our Sintel validation set (Intel Core i7-3930K, 3.2 GHz, 6 cores). The average runtimes are 19.9s for FlowFields and 38.1s for ProbFlowFields, *i.e.* the additional estimation of uncertainties has an overhead of $\sim$1.9x. The best post-hoc uncertainty measure (learned [30]) requires 123.8s on average, thus takes significantly longer than our joint approach.

### 7.4. Application to motion segmentation

We now show the benefit of our uncertainty estimates when optical flow is used as a cue for motion segmentation. Current state-of-the-art methods (*e.g.* [22, 33]) build upon precomputed point trajectories, *i.e.* spatio-temporal curves that describe the individual point motion over extended periods. Ideally, such point trajectories are sampled with a regular density, they are long and reliable [6].

We build on the minimum cost multicut approach of [22] to obtain sparse motion segmentations. We follow Keuper *et al.* and generate point trajectories as in [6]. In a first setup, we use FlowFields [1] to compute forward-backward (FB) flow for all consecutive image pairs. Subsequently, points are sampled on a regular grid in the first frame and tracked as long as *(1)* their FB flows are consistent, and *(2)* the gradient magnitude of the flow is below a threshold. If tracks are stopped, new points are inserted to preserve sampling regularity unless no points can be tracked in a region.

In a second setting, we apply ProbFlowFields using parameters trained on Sintel to compute FB flow along with normalized forward uncertainties $\Psi^{\text{F}}$. Here, we keep the FB condition (1) and use a new condition *(3)* such that we end any track passing through location $(i, j)$ if the pixel uncertainty $\Psi^{\text{F}}_{ij}$ falls below an empirically determined threshold.

Our results in terms of segmentation precision, recall, F-measure, as well as the achieved trajectory density on the FBMS-59 dataset [33] are given in Table 5. For both approaches, sparse trajectories are sampled at 8 pixel distance, and framewise dense segmentations are computed using the variational approach of [32]. Moreover, we compare to the current state-of-the-art [22] on FBMS-59, which is based on Large Displacement Optical Flow (LDOF) [7].

The evaluation of sparse segmentations shows that ProbFlow allows for a higher average point density of more than 1.18% compared to a maximal density of 0.89% with FlowFields [1] or LDOF [7]. Low point density usually indicates flow inconsistencies in homogeneous regions and therefore a more uneven sampling. As can be seen by the increased F-measure on sparse and especially on densified segmentations, the improvement in the trajectory computation directly translates to an improved segmentation quality.

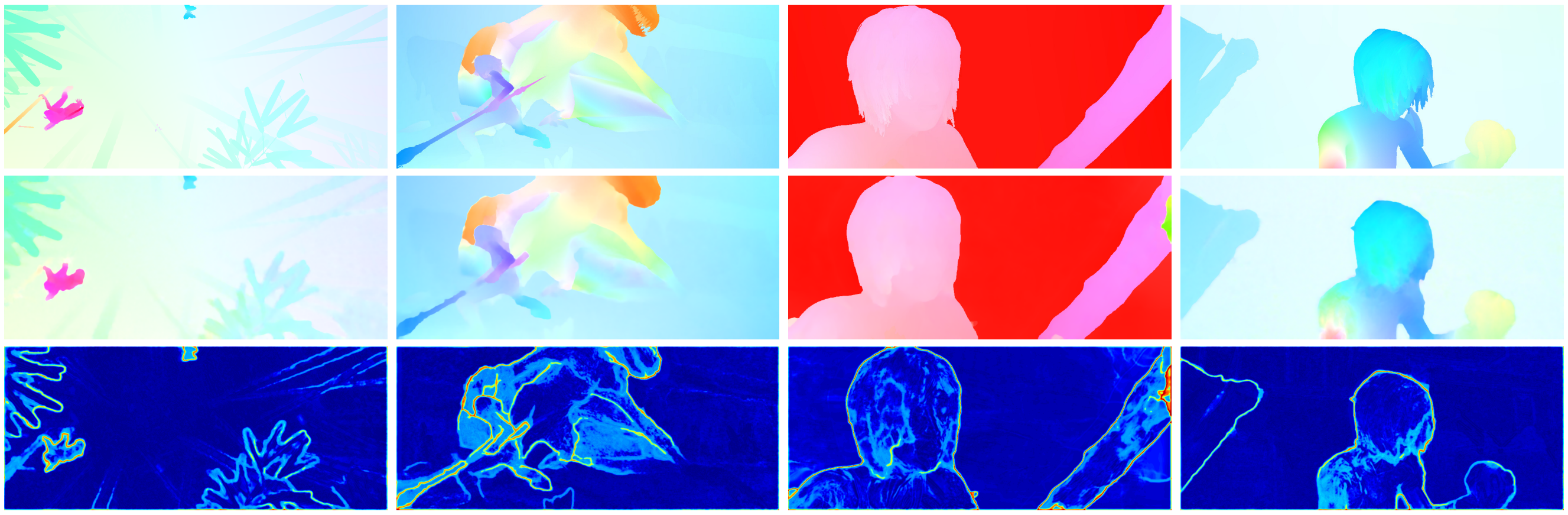

Figure 3. Examples of ground truth (top), flow predictions (middle), and uncertainty estimates (bottom) from ProbFlowFields on Sintel.

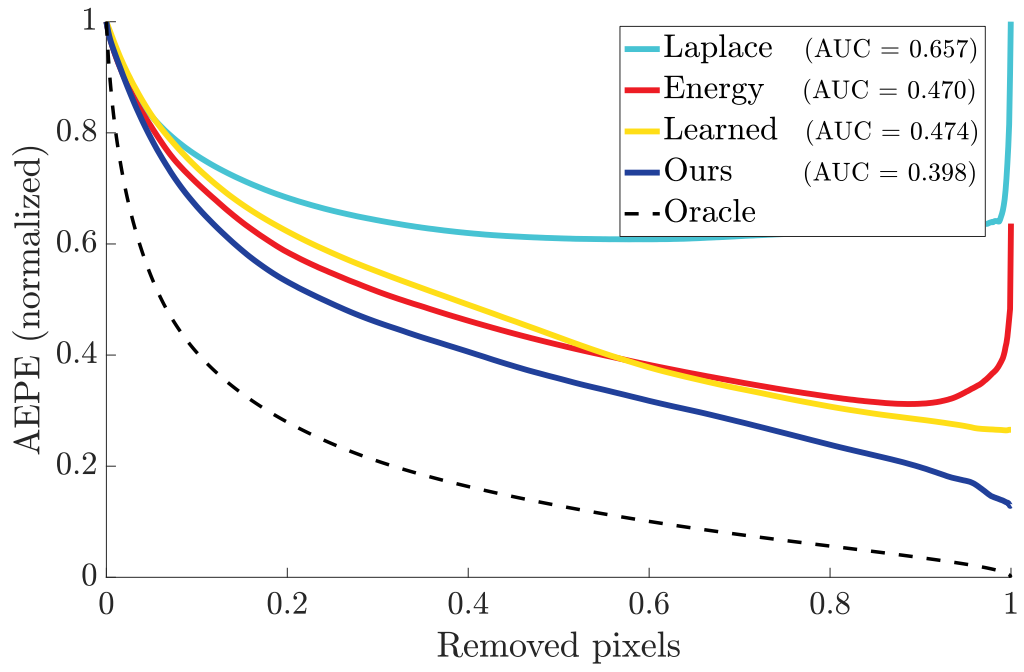

Figure 4. Sparsification plots averaged over Sintel validation.

To illustrate the complementarity of our uncertainties to the FB condition, we perform an experiment using ProbFlowFields estimates and the FB check (1), but not the corresponding uncertainties, *i.e.* omitting (3). In this case, the F-measure of the sparse matches drops significantly. Using FlowFields estimates and the post-hoc measure $\Psi_{\text{Energy}}$, the uncertainties again complement (1). However, the usage of ProbFlowFields yields a higher F-measure, thus showing a clear benefit also in this application.

| Training set (29 seq.) | D | P | R | F |
|---|---|---|---|---|
| LDOF [7] | 0.81% | 86.73% | 73.08% | 79.32% |
| FlowFields (1+2) | 0.83% | 87.19% | 74.33% | 80.25% |
| FlowFields (1) | 1.17% | 85.10% | 71.36% | 77.63% |
| FlowFields + $\Psi_{\text{Energy}}$ | 1.17% | 84.62% | 72.57% | 78.13% |
| ProbFlowFields (1+3) | 1.18% | **87.68%** | **75.13%** | **80.92%** |
| ProbFlowFields (1) | **1.34%** | 84.96% | 72.14% | 78.03% |
| FlowFields [1] dense | 100% | 86.14% | 67.28% | 75.55% |
| ProbFlowFields dense | 100% | **87.00%** | **70.15%** | **77.67%** |
| **Test set (30 seq.)** | **D** | **P** | **R** | **F** |
| LDOF [7] | 0.87% | **87.88%** | 67.70% | 76.48% |
| FlowFields (1+2) | 0.89% | 86.88% | 69.74% | 77.37% |
| ProbFlowFields (1+3) | **1.19%** | 84.99% | **72.83%** | **78.44%** |
| FlowFields [1] dense | 100% | 84.38% | 61.03% | 70.83% |
| ProbFlowFields dense | 100% | **85.41%** | **66.93%** | **75.05%** |

Table 5. Motion segmentation results on the FBMS-59 dataset. We report point density (D), average precision (P), average recall (R), and F-measure. All results are computed for a sparse trajectory sampling at 8 pixel distance with MCe motion segmentation [22]. All results for LDOF are taken from [22].

## 8. Conclusion

To address the issue that optical flow estimates are not equally reliable throughout an image, we introduced ProbFlow – a probabilistic framework for the joint estimation of optical flow and its underlying uncertainty. Starting from conventional energy minimization methods, we derived the posterior distribution and used variational inference to estimate the flow and a model-inherent uncertainty. This is unlike existing uncertainty measures that detect regions of unreliable flow in a post-hoc step. We applied our approach to two different energy formulations on the Middlebury and Sintel benchmarks, where we obtain competitive flow estimates and significantly improved uncertainties. Applying our uncertainty estimates in the context of motion segmentation, we were able to discard erroneous flow estimates and generate highly reliable point trajectories.

**Acknowledgments.** The research leading to these results has received funding from the European Research Council under the European Union's Seventh Framework Programme (FP/2007–2013)/ERC Grant agreement No. 307942. We would like to thank Tobias Plötz and Jochen Gast for helpful discussions.

# ProbFlow: Joint Optical Flow and Uncertainty Estimation
## – Supplemental Material –

Anne S. Wannenwetsch[1] Margret Keuper[2] Stefan Roth[1]
[1]TU Darmstadt [2]University of Mannheim

**Preface.** In this supplemental material we derive update equations for the mean-field inference in Eq. (13) and show a proof for the solution of the Bayesian risk minimization in Eq. (8). We give further implementation details of ProbClassicA and ProbFlowFields, and present an analysis considering different design choices. Finally, we evaluate the performance of additional uncertainty measures and apply, for completeness, ProbFlowFields on the Middlebury benchmark [2].

## A. Mean-field Update Equations

In the following, we show how to derive the mean-field update equations for ProbClassicA. Update equations for ProbFlowFields can be obtained similarly.

**Notation.** Note that strictly speaking the latent variables are given as $\mathbf{h} = (\mathbf{h}_{\gamma,ij,l})_{\gamma,ij}$ with $\gamma \in \{\mathrm{D}, \mathrm{S}_1, \ldots, \mathrm{S}_p, \mathrm{N}_1, \ldots, \mathrm{N}_q\}$, $p = |S(i,j)|$, $q = |N(i,j)|$ as we have separate latent variables for all penalty functions. In the following, we therefore use the notation $\mathbf{h}_{S_e}$, $\mathbf{h}_{N_e}$ and $f_{S_e}(\cdot)$, $f_{N_e}(\cdot)$ for $e \in S(i,j)$ and $e \in N(i,j)$, respectively. The flow vector of the corresponding neighboring pixel is denoted as $\mathbf{y}_e$. Moreover, GSM parameters $\pi_l$ and $\sigma_l$ differ for data, smoothness, and non-local potentials. For better readability, we drop indices D, S, and N that explicitly distinguish between the different GSMs.

**Variational objective.** As shown in Eq. (13), variational parameters $\boldsymbol{\theta}^\star$ are determined to minimize the Kullback-Leibler divergence between posterior $p$ and its approximating distribution $q$, *i.e.*

$$\boldsymbol{\theta}^\star = \arg\min_{\boldsymbol{\theta}} D_{KL}\big(q(\mathbf{y}, \hat{\mathbf{y}}, \mathbf{h}; \boldsymbol{\theta}) \mid p(\mathbf{y}, \hat{\mathbf{y}}, \mathbf{h} \mid I)\big) \tag{22a}$$

$$= \arg\min_{\boldsymbol{\theta}} \underbrace{\mathbb{E}_{q(\mathbf{y},\hat{\mathbf{y}},\mathbf{h};\boldsymbol{\theta})}\big[\log q(\mathbf{y}, \hat{\mathbf{y}}, \mathbf{h}; \boldsymbol{\theta})\big]}_{①\text{, entropy term}} - \underbrace{\mathbb{E}_{q(\mathbf{y},\hat{\mathbf{y}},\mathbf{h};\boldsymbol{\theta})}\big[\log p(\mathbf{y}, \hat{\mathbf{y}}, \mathbf{h} \mid I)\big]}_{②}. \tag{22b}$$

Recall that we defined the variational distribution $q$ in Eq. (11) as

$$q(\mathbf{y}, \hat{\mathbf{y}}, \mathbf{h}; \boldsymbol{\theta}) = \prod_{i,j} \Big[ q\left(\mathbf{y}_{ij}; \boldsymbol{\theta}\right) \cdot q\left(\hat{\mathbf{y}}_{ij}; \boldsymbol{\theta}\right) \cdot \prod_{\gamma} q\left(\mathbf{h}_{\gamma,ij}; \boldsymbol{\theta}\right) \Big]. \tag{23}$$

Then, the entropy term in ① can be split up as follows:

$$\mathbb{E}_{q(\mathbf{y},\hat{\mathbf{y}},\mathbf{h};\boldsymbol{\theta})}\Big[\log q(\mathbf{y}, \hat{\mathbf{y}}, \mathbf{h}; \boldsymbol{\theta})\Big] = \mathbb{E}_{q(\mathbf{y},\hat{\mathbf{y}},\mathbf{h};\boldsymbol{\theta})}\left[\sum_{i,j} \log q\left(\mathbf{y}_{ij}; \boldsymbol{\theta}\right) + \sum_{i,j} \log q\left(\hat{\mathbf{y}}_{ij}; \boldsymbol{\theta}\right) + \sum_{i,j}\sum_{\gamma} \log q\left(\mathbf{h}_{\gamma,ij}; \boldsymbol{\theta}\right)\right] \tag{24a}$$

$$= \sum_{i,j} \mathbb{E}_{q(\mathbf{y}_{ij};\boldsymbol{\theta})} \log q\left(\mathbf{y}_{ij}; \boldsymbol{\theta}\right) + \sum_{i,j} \mathbb{E}_{q(\hat{\mathbf{y}}_{ij};\boldsymbol{\theta})} \log q\left(\hat{\mathbf{y}}_{ij}; \boldsymbol{\theta}\right) + \sum_{i,j}\sum_{\gamma} \mathbb{E}_{q(\mathbf{h}_{\gamma,ij};\boldsymbol{\theta})} \log q\left(\mathbf{h}_{\gamma,ij}; \boldsymbol{\theta}\right). \tag{24b}$$

Using the well-known entropy of a Gaussian and a multinomial distribution, we obtain

$$\mathbb{E}_{q(\mathbf{y},\hat{\mathbf{y}},\mathbf{h};\boldsymbol{\theta})}\Big[\log q(\mathbf{y}, \hat{\mathbf{y}}, \mathbf{h}; \boldsymbol{\theta})\Big] = -\frac{1}{2}\sum_{i,j} \log\big(\det(\boldsymbol{\Sigma}_{ij})\big) - \frac{1}{2}\sum_{i,j} \log\big(\det(\hat{\boldsymbol{\Sigma}}_{ij})\big) + \sum_{i,j}\sum_{\gamma}\sum_{l} k_{\gamma,ij,l}\ \log\ k_{\gamma,ij,l} + \text{const.} \tag{25}$$

In order to evaluate the term ② in Eq. (22b), it is necessary to compute

$$
\begin{aligned}
&\log p(\mathbf{y}, \hat{\mathbf{y}}, \mathbf{h} \mid I) \\
&= \log \Bigg( \frac{1}{Z} \prod_{i,j} \Big[ \prod_l \pi_l^{h_{\mathrm{D},ij,l}} \mathcal{N}\big(f_{\mathrm{D}}(\mathbf{y}_{ij}; I); 0, \sigma_l^2\big)^{h_{\mathrm{D},ij,l}} \Big]^{\lambda_{\mathrm{D}}} \cdot \prod_{e \in S(i,j)} \Big[ \prod_l \pi_l^{h_{\mathrm{S}_e,ij,l}} \mathcal{N}\big(f_{\mathrm{S}_e}(\mathbf{y}_{ij}, \mathbf{y}_e); 0, \sigma_l^2\big)^{h_{\mathrm{S}_e,ij,l}} \Big]^{\lambda_{\mathrm{S}}} \\
&\qquad \cdot \exp\Big[ - f_{\mathrm{C}}(\mathbf{y}_{ij}, \hat{\mathbf{y}}_{ij})^2 \Big]^{\lambda_{\mathrm{C}}} \cdot \prod_{e \in N(i,j)} \Big[ \prod_l \pi_l^{h_{\mathrm{N}_e,ij,l}} \mathcal{N}\big(f_{\mathrm{N}_e}(\hat{\mathbf{y}}_{ij}, \hat{\mathbf{y}}_e); 0, \sigma_l^2\big)^{h_{\mathrm{N}_e,ij,l}} \Big]^{\lambda_{\mathrm{N}}} \Bigg) \qquad \text{(26a)} \\
&= \sum_{i,j} \Bigg[ \lambda_{\mathrm{D}} \sum_l h_{\mathrm{D},ij,l} \Big( \log \pi_l + \log \mathcal{N}\big(f_{\mathrm{D}}(\mathbf{y}_{ij}; I); 0, \sigma_l^2\big) \Big) + \lambda_{\mathrm{S}} \sum_{e \in S(i,j)} \sum_l h_{\mathrm{S}_e,ij,l} \Big( \log \pi_l + \log \mathcal{N}\big(f_{\mathrm{S}_e}(\mathbf{y}_{ij}, \mathbf{y}_e); 0, \sigma_l^2\big) \Big) \\
&\qquad - \lambda_C\, f_{\mathrm{C}}(\mathbf{y}_{ij}, \hat{\mathbf{y}}_{ij})^2 + \lambda_{\mathrm{N}} \sum_{e \in N(i,j)} \sum_l h_{\mathrm{N}_e,ij,l} \Big( \log \pi_l + \log \mathcal{N}\big(f_{\mathrm{N}_e}(\hat{\mathbf{y}}_{ij}, \hat{\mathbf{y}}_e); 0, \sigma_l^2\big) \Big) \Bigg] - \log Z \qquad \text{(26b)} \\
&= \sum_{i,j} \Bigg[ \lambda_{\mathrm{D}} \sum_l h_{\mathrm{D},ij,l} \Big( \log \pi_l - \log \sigma_l - \frac{f_{\mathrm{D}}(\mathbf{y}_{ij}; I)^2}{2\sigma_l^2} \Big) + \lambda_{\mathrm{S}} \sum_{e \in S(i,j)} \sum_l h_{\mathrm{S}_e,ij,l} \Big( \log \pi_l - \log \sigma_l - \frac{f_{\mathrm{S}_e}(\mathbf{y}_{ij}, \mathbf{y}_e)^2}{2\sigma_l^2} \Big) \\
&\qquad - \lambda_{\mathrm{C}}\, f_{\mathrm{C}}(\mathbf{y}_{ij}, \hat{\mathbf{y}}_{ij})^2 + \lambda_{\mathrm{N}} \sum_{e \in N(i,j)} \sum_l h_{\mathrm{N}_e,ij,l} \Big( \log \pi_l - \log \sigma_l - \frac{f_{\mathrm{N}_e}(\hat{\mathbf{y}}_{ij}, \hat{\mathbf{y}}_e)^2}{2\sigma_l^2} \Big) \Bigg] + \text{const}, \qquad \text{(26c)}
\end{aligned}
$$

where we have defined $f_{\mathrm{C}}(\mathbf{y}_{ij}, \hat{\mathbf{y}}_{ij}) = \|\mathbf{y}_{ij} - \hat{\mathbf{y}}_{ij}\|_2$. We now take the expectation over $\mathbf{h}$ and simplify the remaining expectations as

$$
\begin{aligned}
&\mathbb{E}_{q(\mathbf{y},\hat{\mathbf{y}},\mathbf{h};\boldsymbol{\theta})} \log p(\mathbf{y}, \hat{\mathbf{y}}, \mathbf{h} \mid I) \\
&= \mathbb{E}_{q(\mathbf{y},\hat{\mathbf{y}};\boldsymbol{\theta})} \sum_{i,j} \Bigg[ \lambda_{\mathrm{D}} \sum_l k_{\mathrm{D},ij,l} \Big( \log \pi_l - \log \sigma_l - \frac{f_{\mathrm{D}}(\mathbf{y}_{ij}; I)^2}{2\sigma_l^2} \Big) \\
&\qquad + \lambda_{\mathrm{S}} \sum_{e \in S(i,j)} \sum_l k_{\mathrm{S}_e,ij,l} \Big( \log \pi_l - \log \sigma_l - \frac{f_{\mathrm{S}_e}(\mathbf{y}_{ij}, \mathbf{y}_e)^2}{2\sigma_l^2} \Big) \\
&\qquad - \lambda_{\mathrm{C}}\, f_{\mathrm{C}}(\mathbf{y}_{ij}, \hat{\mathbf{y}}_{ij})^2 + \lambda_{\mathrm{N}} \sum_{e \in N(i,j)} \sum_l k_{\mathrm{N}_e,ij,l} \Big( \log \pi_l - \log \sigma_l - \frac{f_{\mathrm{N}_e}(\hat{\mathbf{y}}_{ij}, \hat{\mathbf{y}}_e)^2}{2\sigma_l^2} \Big) \Bigg] + \text{const} \qquad \text{(27a)} \\
&= \sum_{i,j} \Bigg[ \lambda_{\mathrm{D}} \sum_l k_{\mathrm{D},ij,l} \big( \log \pi_l - \log \sigma_l \big) + \lambda_{\mathrm{S}} \sum_{e \in S(i,j)} \sum_l k_{\mathrm{S}_e,ij,l} \big( \log \pi_l - \log \sigma_l \big) \\
&\qquad + \lambda_{\mathrm{N}} \sum_{e \in N(i,j)} \sum_l k_{\mathrm{N}_e,ij,l} \big( \log \pi_l - \log \sigma_l \big) \Bigg] \\
&\quad - \sum_{i,j} \Bigg[ \lambda_{\mathrm{D}} \sum_l \frac{k_{\mathrm{D},ij,l}}{2\sigma_l^2} \underbrace{\mathbb{E}_{q(\mathbf{y};\boldsymbol{\theta})}\, f_{\mathrm{D}}(\mathbf{y}_{ij}; I)^2}_{③,\, g_{\mathrm{D}}} + \lambda_{\mathrm{S}} \sum_{e \in S(i,j)} \sum_l \frac{k_{\mathrm{S}_e,ij,l}}{2\sigma_l^2} \underbrace{\mathbb{E}_{q(\mathbf{y};\boldsymbol{\theta})}\, f_{\mathrm{S}_e}(\mathbf{y}_{ij}, \mathbf{y}_e)^2}_{④,\, g_{\mathrm{S}_e}} \\
&\qquad + \lambda_{\mathrm{C}} \underbrace{\mathbb{E}_{q(\mathbf{y},\hat{\mathbf{y}};\boldsymbol{\theta})}\, f_{\mathrm{C}}(\mathbf{y}_{ij}, \hat{\mathbf{y}}_{ij})^2}_{⑤,\, g_{\mathrm{C}}} + \lambda_{\mathrm{N}} \sum_{e \in N(i,j)} \sum_l \frac{k_{\mathrm{N}_e,ij,l}}{2\sigma_l^2} \underbrace{\mathbb{E}_{q(\hat{\mathbf{y}};\boldsymbol{\theta})}\, f_{\mathrm{N}_e}(\hat{\mathbf{y}}_{ij}, \hat{\mathbf{y}}_e)^2}_{⑥,\, g_{\mathrm{N}_e}} \Bigg] + \text{const}. \qquad \text{(27b)}
\end{aligned}
$$

To solve the expectation value w.r.t. the linearized brightness constancy in ③, we define $a = I_2\left(i + u^0_{ij}, j + v^0_{ij}\right) - I_1\left(i, j\right)$ and $\mathbf{b} = \nabla_2 I_2\left(i + u^0_{ij}, j + v^0_{ij}\right)^{\mathrm{T}}$. Then we have that

$$\begin{aligned} g_{\mathrm{D}}\big(\boldsymbol{\mu}_{ij}, \boldsymbol{\Sigma}_{ij}; I\big) &= \mathbb{E}_{q(\mathbf{y};\boldsymbol{\theta})}\ f_{\mathrm{D}}\big(\mathbf{y}_{ij}; I\big)^2 && \text{(28a)}\\ &= \mathbb{E}_{q(\mathbf{y}_{ij};\boldsymbol{\theta})}\left[\left(a + \mathbf{b}^{\mathrm{T}}\big(\mathbf{y}_{ij} - \mathbf{y}^0_{ij}\big)\right)^2\right] && \text{(28b)}\\ &= \mathbb{E}_{q(\mathbf{y}_{ij};\boldsymbol{\theta})}\left[a^2 + 2a\mathbf{b}^{\mathrm{T}}\big(\mathbf{y}_{ij} - \mathbf{y}^0_{ij}\big) + \big(\mathbf{y}_{ij} - \mathbf{y}^0_{ij}\big)^{\mathrm{T}}\big(\mathbf{b}\mathbf{b}^{\mathrm{T}}\big)\big(\mathbf{y}_{ij} - \mathbf{y}^0_{ij}\big)\right] && \text{(28c)}\\ &= a^2 + 2a\mathbf{b}^{\mathrm{T}}\big(\boldsymbol{\mu}_{ij} - \mathbf{y}^0_{ij}\big) + \big(\boldsymbol{\mu}_{ij} - \mathbf{y}^0_{ij}\big)^{\mathrm{T}}\big(\mathbf{b}\mathbf{b}^{\mathrm{T}}\big)\big(\boldsymbol{\mu}_{ij} - \mathbf{y}^0_{ij}\big) + \mathrm{Tr}\big(\mathbf{b}\mathbf{b}^{\mathrm{T}}\boldsymbol{\Sigma}_{i,j}\big). && \text{(28d)} \end{aligned}$$

We solve the expectation value $g_{\mathrm{S}_e}$ in ④ for an exemplary function $f_{\mathrm{S}_e}\big(\mathbf{y}_{ij}, \mathbf{y}_e\big) = u_{ij} - u_e$. All remaining terms as well as the terms $g_{\mathrm{N}_e}$ in ⑥ can be resolved in the same manner. Using $\mathbf{A}_1 = \begin{pmatrix}1 & 0\\ 0 & 0\end{pmatrix}$, it holds that

$$\begin{aligned} g_{\mathrm{S}_e}\big(\boldsymbol{\mu}_{ij}, \boldsymbol{\Sigma}_{ij}, \boldsymbol{\mu}_e, \boldsymbol{\Sigma}_e\big) &= \mathbb{E}_{q(\mathbf{y};\boldsymbol{\theta})}\ f_{\mathrm{S}_e}\big(\mathbf{y}_{ij}, \mathbf{y}_e\big)^2 && \text{(29a)}\\ &= \mathbb{E}_{q(\mathbf{y}_e;\boldsymbol{\theta})}\ \mathbb{E}_{q(\mathbf{y}_{ij};\boldsymbol{\theta})}\left[\big(\mathbf{y}_{ij} - \mathbf{y}_e\big)^{\mathrm{T}}\mathbf{A}_1\big(\mathbf{y}_{ij} - \mathbf{y}_e\big)\right] && \text{(29b)}\\ &= \mathbb{E}_{q(\mathbf{y}_e;\boldsymbol{\theta})}\left[\big(\boldsymbol{\mu}_{ij} - \mathbf{y}_e\big)^{\mathrm{T}}\mathbf{A}_1\big(\boldsymbol{\mu}_{ij} - \mathbf{y}_e\big) + \mathrm{Tr}\big(\mathbf{A}_1\boldsymbol{\Sigma}_{ij}\big)\right] && \text{(29c)}\\ &= \big(\boldsymbol{\mu}_{ij} - \boldsymbol{\mu}_e\big)^{\mathrm{T}}\mathbf{A}_1\big(\boldsymbol{\mu}_{ij} - \boldsymbol{\mu}_e\big) + \mathrm{Tr}\big(\mathbf{A}_1\boldsymbol{\Sigma}_{ij}\big) + \mathrm{Tr}\big(\mathbf{A}_1\boldsymbol{\Sigma}_e\big) && \text{(29d)}\\ &= \left(\mu^{(1)}_{ij} - \mu^{(1)}_e\right)^2 + \big(\boldsymbol{\Sigma}_{ij}\big)_{1,1} + \big(\boldsymbol{\Sigma}_e\big)_{1,1} && \text{(29e)} \end{aligned}$$

with $\mu^{(1)}_{ij}$ denoting the first (*i.e.*, horizontal) component of the mean flow vector at pixel $(i, j)$. The term $g_{\mathrm{C}}$ in ⑤ can be determined as

$$\begin{aligned} g_{\mathrm{C}}\big(\boldsymbol{\mu}_{ij}, \boldsymbol{\Sigma}_{ij}, \hat{\boldsymbol{\mu}}_{ij}, \hat{\boldsymbol{\Sigma}}_{ij}\big) &= \mathbb{E}_{q(\mathbf{y},\hat{\mathbf{y}};\boldsymbol{\theta})}\ f_{\mathrm{C}}\big(\mathbf{y}_{ij}, \hat{\mathbf{y}}_{ij}\big)^2 && \text{(30a)}\\ &= \mathbb{E}_{q(\mathbf{y}_{ij};\boldsymbol{\theta})}\ \mathbb{E}_{q(\hat{\mathbf{y}}_{ij};\boldsymbol{\theta})}\left[\big(\mathbf{y}_{ij} - \hat{\mathbf{y}}_{ij}\big)^{\mathrm{T}}\big(\mathbf{y}_{ij} - \hat{\mathbf{y}}_{ij}\big)\right] && \text{(30b)}\\ &= \mathbb{E}_{q(\hat{\mathbf{y}}_{ij};\boldsymbol{\theta})}\left[\big(\boldsymbol{\mu}_{ij} - \hat{\mathbf{y}}_{ij}\big)^{\mathrm{T}}\big(\boldsymbol{\mu}_{ij} - \hat{\mathbf{y}}_{ij}\big) + \mathrm{Tr}\big(\boldsymbol{\Sigma}_{ij}\big)\right] && \text{(30c)}\\ &= \big(\boldsymbol{\mu}_{ij} - \hat{\boldsymbol{\mu}}_{ij}\big)^{\mathrm{T}}\big(\boldsymbol{\mu}_{ij} - \hat{\boldsymbol{\mu}}_{ij}\big) + \mathrm{Tr}\big(\boldsymbol{\Sigma}_{ij}\big) + \mathrm{Tr}\big(\hat{\boldsymbol{\Sigma}}_{ij}\big). && \text{(30d)} \end{aligned}$$

**Update equations.** To obtain update equations, we compute the derivative of the KL divergence in Eq. (22b), set it to zero, and solve for the desired variable. Please note that update equations for boundary pixels may slightly differ from the ones shown below. From now on, spatial derivatives of $I$ are denoted as $I_x$ and $I_y$, the temporal derivative is given as $I_t$. Moreover, $\mathrm{diag}(\cdot)$ represents a diagonal matrix and we define vectors $\mathbf{K}_\gamma = \left(\sum_l \frac{\mathbf{k}_{\gamma,ij,l}}{\sigma_l^2}\right)_{ij}$.

As the update of each mean flow estimate $\boldsymbol{\mu}_{ij}$ depends on other entries of $\boldsymbol{\mu}$, it is desirable to jointly solve for all components of the flow field. Therefore, $\boldsymbol{\mu}$ is obtained as the solution of a linear equation system, *c.f.* [50, 37], such that

$$\mathbf{A}\mathbf{x} = \mathbf{b}, \quad \mathbf{x} = \left(\mu^{(1)}_{11}, \dots, \mu^{(1)}_{nm}, \mu^{(2)}_{11}, \dots, \mu^{(2)}_{nm}\right)^{\mathrm{T}}, \quad \mathbf{A} = \mathbf{A}_{\mathrm{D}} + \mathbf{A}_{\mathrm{S}} + \mathbf{A}_{\mathrm{C}}, \quad \mathbf{b} = \mathbf{b}_{\mathrm{D}} + \mathbf{b}_{\mathrm{S}} + \mathbf{b}_{\mathrm{C}}. \tag{31}$$

The components of the linear equation system are determined as

$$\begin{aligned} \mathbf{A}_{\mathrm{D}} &= \lambda_{\mathrm{D}}\begin{pmatrix}\mathrm{diag}\left(\mathbf{K}_{\mathrm{D}}\right)\mathrm{diag}\left(I_x^2\right) & \mathrm{diag}\left(\mathbf{K}_{\mathrm{D}}\right)\mathrm{diag}\left(I_x \cdot I_y\right)\\ \mathrm{diag}\left(\mathbf{K}_{\mathrm{D}}\right)\mathrm{diag}\left(I_x \cdot I_y\right) & \mathrm{diag}\left(\mathbf{K}_{\mathrm{D}}\right)\mathrm{diag}\left(I_y^2\right)\end{pmatrix}, && \text{(32a)}\\ \mathbf{A}_{\mathrm{S}} &= \lambda_{\mathrm{S}}\begin{pmatrix}\mathbf{F}_1^{\mathrm{T}}\mathrm{diag}\left(\mathbf{K}_{\mathrm{S}_1}\right)\mathbf{F}_1 + \mathbf{F}_2^{\mathrm{T}}\mathrm{diag}\left(\mathbf{K}_{\mathrm{S}_2}\right)\mathbf{F}_2 & \mathbf{0}\\ \mathbf{0} & \mathbf{F}_1^{\mathrm{T}}\mathrm{diag}\left(\mathbf{K}_{\mathrm{S}_3}\right)\mathbf{F}_1 + \mathbf{F}_2^{\mathrm{T}}\mathrm{diag}\left(\mathbf{K}_{\mathrm{S}_4}\right)\mathbf{F}_2\end{pmatrix}, && \text{(32b)}\\ \mathbf{A}_{\mathrm{C}} &= 2\lambda_{\mathrm{C}}\mathbf{I}, && \text{(32c)}\\ \mathbf{b}_{\mathrm{D}} &= \mathbf{A}_D\mathbf{y}_0 - \lambda_{\mathrm{D}}\begin{pmatrix}\mathrm{diag}\left(\mathbf{K}_{\mathrm{D}}\right)\mathrm{diag}\left(I_x \cdot I_t\right)\mathbf{1}\\ \mathrm{diag}\left(\mathbf{K}_{\mathrm{D}}\right)\mathrm{diag}\left(I_y \cdot I_t\right)\mathbf{1}\end{pmatrix},\ \mathbf{b}_{\mathrm{S}} = \mathbf{0},\ \mathbf{b}_{\mathrm{C}} = 2\lambda_{\mathrm{C}}\hat{\boldsymbol{\mu}}. && \text{(32d)} \end{aligned}$$

Here, $\mathbf{F}_1$ and $\mathbf{F}_2$ represent filter matrices corresponding to the derivative filters $\mathbf{H}_1 = \left[1, -1\right]^\mathrm{T}$ and $\mathbf{H}_2 = \left[1, -1\right]$, which are used in $f_{\mathrm{S}_e}\left(\mathbf{y}_{ij}, \mathbf{y}_e\right)$, *c.f.* [37]. $\mathbf{I}$ is the identity matrix and $\mathbf{0}$ is a matrix of all zeros.

When updating the auxiliary flow means $\hat{\boldsymbol{\mu}}$, a $5 \times 5$ neighborhood has to be considered. Therefore, a joint update of all estimates is computationally expensive and we follow [43] assuming fixed values for neighboring pixels, *i.e.*

$$\hat{\boldsymbol{\mu}}_{ij,t} = \begin{pmatrix} \hat{\mu}^{(1)}_{ij,t} \\ \hat{\mu}^{(2)}_{ij,t} \end{pmatrix}, \quad \hat{\mu}^{(k)}_{ij,t} = \frac{2\lambda_\mathrm{C}\, \mu^{(k)}_{ij} + \lambda_\mathrm{N} \sum_{e \in N^k(i,j)} \left(\mathbf{K}_{N^k_e}\right)_{ij,t-1} \cdot \hat{\mu}^{(k)}_{e,t-1}}{2\lambda_\mathrm{C} + \lambda_\mathrm{N} \sum_{e \in N^k(i,j)} \left(\mathbf{K}_{N^k_e}\right)_{ij,t-1}}. \tag{33}$$

Here, $N^k(i,j)$ represents the set of neighbors in terms of the $k^{\text{th}}$ optical flow component.

For the flow variances $\boldsymbol{\Sigma} = \begin{pmatrix} \Sigma_1 & 0 \\ 0 & \Sigma_2 \end{pmatrix}$ we derive a closed-form update dependent only on the latent variables $\mathbf{k}$ with

$$\Sigma_1 = \left(\lambda_\mathrm{D} \operatorname{diag}\left(I_x^2\right) \cdot \mathbf{K}_\mathrm{D} + \lambda_\mathrm{S} \left[\operatorname{abs}\left(\mathbf{F}_1^\mathrm{T}\right) \cdot \mathbf{K}_{\mathrm{S}_1} + \operatorname{abs}\left(\mathbf{F}_2^\mathrm{T}\right) \cdot \mathbf{K}_{\mathrm{S}_2}\right] + 2\lambda_\mathrm{C}\right)^{-1} \tag{34a}$$

and $$\Sigma_2 = \left(\lambda_\mathrm{D} \operatorname{diag}\left(I_y^2\right) \cdot \mathbf{K}_\mathrm{D} + \lambda_\mathrm{S} \left[\operatorname{abs}\left(\mathbf{F}_1^\mathrm{T}\right) \cdot \mathbf{K}_{\mathrm{S}_3} + \operatorname{abs}\left(\mathbf{F}_2^\mathrm{T}\right) \cdot \mathbf{K}_{\mathrm{S}_4}\right] + 2\lambda_\mathrm{C}\right)^{-1}, \tag{34b}$$

where the absolute value function $\operatorname{abs}(\cdot)$ is applied element-wise.

We assume fixed neighboring values also for the update of the auxiliary flow variances $\hat{\boldsymbol{\Sigma}}$, and obtain

$$\hat{\boldsymbol{\Sigma}}_{ij,t} = \begin{pmatrix} \hat{\Sigma}^{(1)}_{ij,t} & 0 \\ 0 & \hat{\Sigma}^{(2)}_{ij,t} \end{pmatrix}, \quad \hat{\Sigma}^{(k)}_{ij,t} = \frac{1}{2\lambda_\mathrm{C} + \lambda_\mathrm{N} \sum_{e \in N^k(i,j)} \left(\mathbf{K}_{N^k_e}\right)_{ij,t-1}}. \tag{35}$$

To derive an update equation for a latent variable $\mathbf{k}_{\gamma,ij}$, we need to consider a Lagrangian function including the KL divergence in Eq. (22b) as well as the constraint $\sum_l k_{\gamma,ij,l} = 1$. Solving the resulting linear equation system analytically gives us, *e.g.*,

$$k_{\mathrm{D},ij,l} = \left(\frac{\pi_l}{\sigma_l}\right)^{\lambda_\mathrm{D}} \exp\left[-\lambda_\mathrm{D} \frac{g_\mathrm{D}\left(\boldsymbol{\mu}_{ij}, \boldsymbol{\Sigma}_{ij}; I\right)}{2\sigma_l^2}\right] \cdot Z_{\mathrm{D},ij} \tag{36a}$$

with $$Z_{\mathrm{D},ij} = \left(\sum_{l=1}^{L} \left(\frac{\pi_l}{\sigma_l}\right)^{\lambda_\mathrm{D}} \exp\left[-\lambda_\mathrm{D} \frac{g_\mathrm{D}\left(\boldsymbol{\mu}_{ij}, \boldsymbol{\Sigma}_{ij}; I\right)}{2\sigma_l^2}\right]\right)^{-1} \tag{36b}$$

for the latent variables of the data term using the expectation values $g_\mathrm{D}\left(\boldsymbol{\mu}_{ij}, \boldsymbol{\Sigma}_{ij}; I\right)$ as derived in Eq. (28d). Update equations for the remaining latent variables are derived similarly.

## B. Bayesian Risk Minimization

We aim to show that the solution of the Bayesian risk minimization in Eq. (8) is given as $\mathbf{y}^\star_{ij} = \boldsymbol{\mu}_{ij}$ when replacing the posterior $p$ with its approximating distribution $q$ and using the Average End-Point Error (AEPE) as a loss function.

Recall that the AEPE is defined as $l\left(\mathbf{y}, \tilde{\mathbf{y}}\right) = \sum_{i,j} \ell\left(\mathbf{y}_{ij}, \tilde{\mathbf{y}}_{ij}\right) = \sum_{i,j} \|\mathbf{y}_{ij} - \tilde{\mathbf{y}}_{ij}\|_2$ with

$$\nabla_2\, \ell\left(\mathbf{a} - \mathbf{x}, \tilde{\mathbf{x}}\right) = \left(\mathbf{a} - \mathbf{x} - \tilde{\mathbf{x}}\right) / \|\mathbf{a} - \mathbf{x} - \tilde{\mathbf{x}}\|_2 \tag{37a}$$
$$= -\left(\mathbf{x} - \left(\mathbf{a} - \tilde{\mathbf{x}}\right)\right) / \|\mathbf{x} - \left(\mathbf{a} - \tilde{\mathbf{x}}\right)\|_2 \tag{37b}$$
$$= -\nabla_2\, \ell\left(\mathbf{x}, \mathbf{a} - \tilde{\mathbf{x}}\right) \tag{37c}$$

for arbitrary $\mathbf{a} \in \mathbb{R}^2$. W.l.o.g. we minimize the expected risk of $l(\mathbf{y}, \tilde{\mathbf{y}})$ and therefore set $f(\tilde{\mathbf{y}}) = \mathbb{E}_{q(\mathbf{y}, \hat{\mathbf{y}}, \mathbf{h}; \boldsymbol{\theta})}\left[l(\mathbf{y}, \tilde{\mathbf{y}})\right]$. Note that we omit the variational parameters $\boldsymbol{\theta}$ in the following for brevity. Using the properties of $q$, we obtain

$$f\left(\tilde{\mathbf{y}}\right) = \int_{\mathcal{Y}} \int_{\hat{\mathcal{Y}}} \sum_{\mathcal{H}} q(\mathbf{y}, \hat{\mathbf{y}}, \mathbf{h}) \cdot l\left(\mathbf{y}, \tilde{\mathbf{y}}\right) d\mathbf{y}\; d\hat{\mathbf{y}} \tag{38a}$$
$$\overset{\text{q fac.}}{=} \int_{\mathcal{Y}} q(\mathbf{y}) \cdot l\left(\mathbf{y}, \tilde{\mathbf{y}}\right) d\mathbf{y} \tag{38b}$$
$$= \sum_{i,j} \underbrace{\int_{\mathbb{R}^2} q(\mathbf{y}_{ij}) \cdot \ell\left(\mathbf{y}_{ij}, \tilde{\mathbf{y}}_{ij}\right) d\mathbf{y}_{ij}}_{=: f_{ij}(\tilde{\mathbf{y}}_{ij})}. \tag{38c}$$

For fixed $\mathbf{y}_{ij} \in \mathbb{R}^2$, the function $q(\mathbf{y}_{ij}) \cdot \ell(\mathbf{y}_{ij}, \tilde{\mathbf{y}}_{ij})$ is convex in $\tilde{\mathbf{y}}_{ij}$. Therefore, the objective $f(\tilde{\mathbf{y}})$ is convex in $\tilde{\mathbf{y}}$ and the Bayesian risk minimization has a unique solution given by

$$\mathbf{y}_{ij} = \arg\min_{\tilde{\mathbf{y}}_{ij}} f_{ij}(\tilde{\mathbf{y}}_{ij}). \tag{39}$$

It only remains to be shown that $\nabla_2 f_{ij}(\tilde{\mathbf{y}}_{ij}) = 0$ holds for $\tilde{\mathbf{y}}_{ij} = \boldsymbol{\mu}_{ij}$. Setting $\tilde{\mathbf{y}}_{ij} = \boldsymbol{\mu}_{ij}$ we obtain

$$\begin{aligned}
\int_{-\infty}^{\boldsymbol{\mu}_{ij}} q(\boldsymbol{\tau}) \cdot \nabla_2\, \ell\left(\boldsymbol{\tau}, \boldsymbol{\mu}_{ij}\right) d\boldsymbol{\tau} &\overset{(\mathbf{z}_1 = \boldsymbol{\tau} - \boldsymbol{\mu}_{ij})}{=} \int_{-\infty}^{\mathbf{0}} q(\boldsymbol{\mu}_{ij} + \mathbf{z}_1) \cdot \nabla_2\, \ell\left(\boldsymbol{\mu}_{ij} + \mathbf{z}_1, \boldsymbol{\mu}_{ij}\right) d\mathbf{z}_1 && \text{(40a)}\\
&\overset{q \text{ sym.}}{=} \int_{-\infty}^{\mathbf{0}} q(\boldsymbol{\mu}_{ij} - \mathbf{z}_1) \cdot \nabla_2\, \ell\left(\boldsymbol{\mu}_{ij} + \mathbf{z}_1, \boldsymbol{\mu}_{ij}\right) d\mathbf{z}_1 && \text{(40b)}\\
&\overset{(\mathbf{z}_2 = \boldsymbol{\mu}_{ij} - \mathbf{z}_1)}{=} \int_{\boldsymbol{\mu}_{ij}}^{\infty} q(\mathbf{z}_2) \cdot \nabla_2\, \ell\left(2\boldsymbol{\mu}_{ij} - \mathbf{z}_2, \boldsymbol{\mu}_{ij}\right) d\mathbf{z}_2 && \text{(40c)}\\
&\overset{(37a)-(37c)}{=} -\int_{\boldsymbol{\mu}_{ij}}^{\infty} q(\mathbf{z}_2) \cdot \nabla_2\, \ell\left(\mathbf{z}_2, \boldsymbol{\mu}_{ij}\right) d\mathbf{z}_2 && \text{(40d)}
\end{aligned}$$

and finally

$$\begin{aligned}
\nabla_2\, f_{ij}\left(\boldsymbol{\mu}_{ij}\right) &= \int_{\mathbb{R}^2} q(\boldsymbol{\tau}) \cdot \nabla_2\, \ell\left(\boldsymbol{\tau}, \boldsymbol{\mu}_{ij}\right) d\boldsymbol{\tau} && \text{(41a)}\\
&= \int_{-\infty}^{\boldsymbol{\mu}_{ij}} q(\boldsymbol{\tau}) \cdot \nabla_2\, \ell\left(\boldsymbol{\tau}, \boldsymbol{\mu}_{ij}\right) d\boldsymbol{\tau} + \int_{\boldsymbol{\mu}_{ij}}^{\infty} q(\boldsymbol{\tau}) \cdot \nabla_2\, \ell\left(\boldsymbol{\tau}, \boldsymbol{\mu}_{ij}\right) d\boldsymbol{\tau} \\
&\overset{(40d)}{=} -\int_{\boldsymbol{\mu}_{ij}}^{\infty} q(\boldsymbol{\tau}) \cdot \nabla_2\, \ell\left(\boldsymbol{\tau}, \boldsymbol{\mu}_{ij}\right) d\boldsymbol{\tau} + \int_{\boldsymbol{\mu}_{ij}}^{\infty} q(\boldsymbol{\tau}) \cdot \nabla_2\, \ell\left(\boldsymbol{\tau}, \boldsymbol{\mu}_{ij}\right) d\boldsymbol{\tau} && \text{(41b)}\\
&= 0. && \text{(41c)}
\end{aligned}$$

## C. Implementation Details

In this section, we present our design choices following the best-practices of energy-based optical flow techniques, and give an analysis evaluating the influence of the specifics. Moreover, we give details of our post-processing approach using the fast bilateral solver [4].

### C.1. ProbClassicA

In our ProbClassicA algorithm, we perform three steps of graduated non-convexity and apply coarse-to-fine estimation with 10 warping steps per layer. As in [43], we restrict the flow update to an absolute value of 1 and pre-process the images using a structure-texture decomposition. Spline-based cubic interpolation as well as an averaging of image gradients $\nabla_2 I_1$ and $\nabla_2 I_2$ are applied. During the inference, the variable sets $\{\boldsymbol{\mu}, \boldsymbol{\Sigma}, \mathbf{k}\}$ and $\{\hat{\boldsymbol{\mu}}, \hat{\boldsymbol{\Sigma}}, \hat{\mathbf{k}}\}$ are updated in an alternating way. As an inner update step, we apply five iterations of the block-coordinate descent scheme on $\boldsymbol{\mu}$, $\boldsymbol{\Sigma}$ and $\mathbf{k}$. For the set $\{\hat{\boldsymbol{\mu}}, \hat{\boldsymbol{\Sigma}}, \hat{\mathbf{k}}\}$, a number of three inner updates performs better.

### C.2. ProbFlowFields

For ProbFlowFields, we follow [36] and pre-smooth images using a Gaussian kernel of size $9 \times 9$ with $\sigma = 1.1$. For warping, we apply bilinear interpolation and averaged image derivatives. Moreover, we perform five warping steps, each with five iterations of our block-coordinate descent scheme. We follow Revaud *et al.* and compute optical flow updates with 30 iterations of successive over relaxation, which performs noticeably faster than the solver used in [43].

### C.3. Evaluation of design choices

Table 6 summarizes results of AEPE, AUC, and CC on the Middlebury and Sintel benchmarks using varying setups of ProbClassicA and ProbFlowFields. In a first step, we evaluate a setting for ProbClassicA in which parameters $\lambda_{\mathrm{D}}$, $\lambda_{\mathrm{S}}$, and $\lambda_{\mathrm{N}}$ are determined by having the Bayesian optimization [42] consider only the AEPE or only the AUC instead of the $\mathrm{F}_1$-score proposed in Eq. (20). In both cases, we observe that the performance w.r.t. the evaluation metric that is not considered during the Bayesian optimization drops significantly. This highlights the importance of the $\text{F}_1$-score to balance the accuracy of flow and uncertainty estimates. Moreover, we show that the AEPE as well as the performance of the uncertainty measure is clearly inferior if no additional nonlocal term is applied ($E_{\text{N}} = E_{\text{C}} = 0$). When using ProbClassicA without structure-texture decomposition as pre-processing, we surprisingly obtain improved results for the AEPE ($2\%$) as well as the AUC ($5\%$). This is in contrast to energy minimization, where this pre-processing helps [43]. For fairness of comparison to the underlying energy minimization approach, we continue to use a structure-texture decomposition.

| ProbClassicA Middlebury | AEPE | rel. chg. | AUC | rel. chg. | CC | rel. chg. |
|---|---|---|---|---|---|---|
| Baseline | 0.296 | – | 0.466 | – | 0.374 | – |
| Bayesian optim. w.r.t. AEPE only | 0.290 | **-0.02** | 0.471 | 0.01 | 0.351 | **0.06** |
| Bayesian optim. w.r.t. AUC only | 0.312 | **0.05** | 0.436 | **-0.06** | 0.451 | **-0.21** |
| $E_{\text{N}} = E_{\text{C}} = 0$ | 0.411 | **0.39** | 0.889 | **0.91** | 0.125 | **0.67** |
| No structure-texture decomposition | 0.290 | **-0.02** | 0.445 | **-0.05** | 0.361 | **0.03** |
| **ProbFlowFields Sintel validation** | **AEPE** | **rel. chg.** | **AUC** | **rel. chg.** | **CC** | **rel. chg.** |
| Baseline | 3.127 | – | 0.398 | – | 0.563 | – |
| Bayesian optim. w.r.t. AEPE only | 3.128 | <0.01 | 0.475 | **0.19** | 0.407 | **0.28** |
| Bayesian optim. w.r.t. AUC only | 3.219 | **0.03** | 0.381 | **-0.04** | 0.644 | **-0.14** |
| Spatially constant $\lambda_{\text{S}}$ | 3.127 | 0.00 | 0.400 | <0.01 | 0.562 | <0.01 |
| $\theta_{ij}^{r} = 1$ | 3.125 | >-0.01 | 0.396 | >-0.01 | 0.548 | **0.03** |
| No gradient averaging | 3.135 | <0.01 | 0.398 | 0.00 | 0.557 | 0.01 |
| No Gaussian smoothing | 3.135 | <0.01 | 0.441 | **0.11** | 0.497 | **0.12** |
| 10 warping steps | 3.123 | >-0.01 | 0.421 | **0.06** | 0.538 | **0.04** |

Table 6. Analysis of several design choices for ProbClassicA on Middlebury and ProbFlowFields on the Sintel validation set. Bold entries denote strong deviations from the baseline.

Considering ProbFlowFields, we observe the same behavior as for ProbClassicA when Bayesian optimization is carried out only with respect to one of the evaluation metrics. Note that the parameter setting obtained by a Bayesian optimization w.r.t. to the AEPE performs better than the baseline on the training set even though no improvement of the AEPE is visible on the validation set. The usage of a spatially constant trade-off parameter $\lambda_{\text{S}}$, turning off the normalization of the spatial derivatives ($\theta_{ij}^{r} = 1$, *c.f.* Eqs. (17) and (18)), and not averaging the image gradients, respectively, only lead to minor changes. When no Gaussian smoothing is applied for image pre-processing, a clear effect on the AUC as well as the CC can be observed whereas the AEPE is only slightly changed. Finally, the application of 10 warping steps only results in small improvements of the AEPE and even decreases the performance of the uncertainty measure. This justifies the usage of a reduced number of 5 steps to save computational time.

### C.4. Post-processing using the fast bilateral solver

As described in Sec. 7.3, we apply the fast bilateral solver [4] on top of ProbFlowFields in order to illustrate the benefits of uncertainty predictions for a further improvement of the flow estimates. In doing so, we normalize the estimated uncertainties with a sigmoid function and invert the values to obtain the confidences required by the fast bilateral solver. A Bayesian optimization [42] is performed on our Sintel training set to obtain appropriate sigmoid parameters as well as a suitable trade-off parameter for the fast bilateral solver. See Fig. 5 for a screenshot of the private Sintel benchmark table showing results after post-processing (ProbFlowFields + BS). For the reported baseline, we process the estimates of ProbFlowFields assuming a uniform confidence of $0.5$.

## D. Additional Uncertainty Measures

In the following, we evaluate several additional uncertainty measures on the Middlebury as well as the Sintel benchmark. Haußecker and Spies [20] introduce three confidence measures based on the spatio-temporal structure tensor

$$\mathbf{S} = G(\tilde{\sigma}) * \left[(\nabla_3 I)(\nabla_3 I)^{\mathrm{T}}\right] \qquad \text{with} \qquad \nabla_3 I = (I_x, I_y, I_t)^{\mathrm{T}}, \tag{42}$$

where $I_x$ and $I_y$ denote the spatial image derivatives computed with central differences and $I_t$ is the temporal difference between $I_1$ and $I_2$. Following [30], we smooth the derivatives with a Gaussian filter $G(\tilde{\sigma})$ of size $7 \times 7$ and a standard deviation $\tilde{\sigma} = 2$. In [20], eigenvalues $\lambda_1$, $\lambda_2$, and $\lambda_3$ of $\mathbf{S}$ are computed such that $\lambda_1 \geq \lambda_2 \geq \lambda_3$. Uncertainty measures are then obtained as

$$\Psi_{\text{Ct}} = -\left(\frac{\lambda_1 - \lambda_3}{\lambda_1 + \lambda_3}\right)^2, \quad \Psi_{\text{Cs}} = -\left(\frac{\lambda_1 - \lambda_2}{\lambda_1 + \lambda_2}\right)^2, \quad \text{and} \quad \Psi_{\text{Cc}} = \Psi_{\text{Ct}} - \Psi_{\text{Cs}}. \tag{43}$$

| | EPE all | EPE matched | EPE unmatched | d0-10 | d10-60 | d60-140 | s0-10 | s10-40 | s40+ | |
|---|---|---|---|---|---|---|---|---|---|---|
| GroundTruth [1] | 0.000 | 0.000 | 0.000 | 0.000 | 0.000 | 0.000 | 0.000 | 0.000 | 0.000 | Visualize Results |
| DCFlow [2] | 5.119 | 2.283 | 28.228 | 4.665 | 2.108 | 1.440 | 1.052 | 3.434 | 29.351 | Visualize Results |
| FlowFieldsCNN [3] | 5.363 | 2.303 | 30.313 | 4.718 | 2.020 | 1.399 | 1.032 | 3.065 | 32.422 | Visualize Results |
| MR-Flow [4] | 5.376 | 2.818 | 26.235 | 5.109 | 2.395 | 1.755 | 0.908 | 3.443 | 32.221 | Visualize Results |
| FTFlow [5] | 5.390 | 2.268 | 30.841 | 4.513 | 1.964 | 1.366 | 1.046 | 3.322 | 31.936 | Visualize Results |
| S2F-IF [6] | 5.417 | 2.549 | 28.795 | 4.745 | 2.198 | 1.712 | 1.157 | 3.468 | 31.262 | Visualize Results |
| InterpoNet_ff [7] | 5.535 | 2.372 | 31.296 | 4.720 | 2.018 | 1.532 | 1.064 | 3.496 | 32.633 | Visualize Results |
| RegionalFF [8] | 5.562 | 2.595 | 29.741 | 4.921 | 2.393 | 1.639 | 1.122 | 3.477 | 32.625 | Visualize Results |
| PGM-C [9] | 5.591 | 2.672 | 29.389 | 4.975 | 2.340 | 1.791 | 1.057 | 3.421 | 33.339 | Visualize Results |
| RicFlow [10] | 5.620 | 2.765 | 28.907 | 5.146 | 2.366 | 1.679 | 1.088 | 3.364 | 33.573 | Visualize Results |
| InterpoNet_cpm [11] | 5.627 | 2.594 | 30.344 | 4.975 | 2.213 | 1.640 | 1.042 | 3.575 | 33.321 | Visualize Results |
| ProbFlowFields+BS [12] | 5.628 | 2.543 | 30.773 | 4.680 | 2.169 | 1.683 | 1.086 | 3.538 | 33.210 | Visualize Results |
| CPM_AUG [13] | 5.645 | 2.737 | 29.362 | 4.707 | 2.150 | 1.918 | 1.087 | 3.306 | 33.925 | Visualize Results |
| ProbFlowFields [14] | 5.696 | 2.545 | 31.371 | 4.696 | 2.150 | 1.686 | 1.146 | 3.658 | 33.188 | Visualize Results |

Figure 5. Screenshot of private Sintel table (final) showing results for ProbFlowFields and ProbFlowFields + BS (status as of July 2017).

| Uncertainty measure | AUC | rel. chg. | CC | rel. chg. |
|---|---|---|---|---|
| Ct [20] | 1.058 | 1.27 | -0.106 | 1.28 |
| Cs [20] | 1.014 | 1.18 | -0.057 | 1.15 |
| Cc [20] | 0.967 | 1.08 | -0.022 | 1.06 |
| Ev3 [27] | 0.989 | 1.12 | 0.058 | 0.84 |
| Noise | 0.512 | 0.10 | 0.286 | 0.24 |
| ProbClassicA *(ours)* | **0.466** | **0.00** | **0.374** | **0.00** |
| Oracle | *0.255* | – | *1.000* | – |

Table 7. Area under curve (AUC), Spearman's rank correlation coefficient (CC), and relative change (rel. chg.) in comparison to the our uncertainty measure on the Middlebury dataset.

| Uncertainty measure | AUC | rel. chg. | CC | rel. chg. |
|---|---|---|---|---|
| Ct [20] | 1.130 | 1.84 | -0.128 | 1.23 |
| Cs [20] | 1.154 | 1.90 | -0.149 | 1.26 |
| Cc [20] | 0.915 | 1.30 | 0.129 | 0.77 |
| Ev3 [27] | 1.024 | 1.57 | -0.030 | 1.05 |
| Noise | 0.512 | 0.29 | 0.382 | 0.32 |
| ProbFlowFields *(ours)* | **0.398** | **0.00** | **0.563** | **0.00** |
| Oracle | *0.182* | – | *1.000* | – |

Table 8. Area under curve (AUC), Spearman's rank correlation coefficient (CC), and relative change (rel. chg.) in comparison to our uncertainty measure on a Sintel benchmark validation set.

Moreover, we evaluate a baseline uncertainty measure as used in [27] defined as $\Psi_{\text{Ev3}} = -\lambda_3$.

Finally, we compare to a sampling-based measure similar to the idea of Kybic and Nieuwenhuis [27]. That is, we estimate the uncertainty as the variance of the optical flow estimates resulting from small, random perturbations of the input data. Specifically, we apply zero-mean Gaussian noise on the input images and determine appropriate values for the variance of the noise on the training set. The uncertainty measure is then obtained as $\Psi_{\text{Noise}} = \sqrt{\sigma_u^2 + \sigma_v^2}$ with $\sigma_u$ and $\sigma_v$ denoting the standard derivation of the horizontal and vertical flow estimates per pixel.

As can be seen in Tables 7 and 8, all measures based on the structure tensor perform considerably worse than our proposed uncertainty measure. $\Psi_{\text{Cc}}$ and $\Psi_{\text{Ev3}}$ lead to more meaningful uncertainties than the two remaining approaches on both datasets, but perform similar to the simple gradient-based measure [3]. The noise uncertainty – especially on the Middlebury dataset – performs comparably to $\Psi_{\text{Energy}}$ and $\Psi_{\text{Learned}}$. However, our ProbFlow approach clearly leads to superior results.

## E. ProbFlowFields on Middlebury

For completeness, we report the results of ProbFlowFields on Middlebury. To reproduce the Middlebury results shown in [1] we applied the default settings of the EpicFlow interpolation. Moreover, we use GSM potentials trained on the Sintel

| Method | training | | test | |
|---|---|---|---|---|
| | AEPE | rel. chg. | AEPE | rel. chg. |
| Initialization | 0.307 | 0.38 | – | – |
| FlowFields [1] | 0.240 | 0.08 | 0.331$^{\dagger}$ | 0.10 |
| FieldsFields$^{\star}$ | 0.230 | 0.04 | – | – |
| ProbFlowFields *(ours)* | **0.222** | **0.00** | **0.301** | **0.00** |

Table 9. Average end-point error (AEPE) and relative change (rel. chg.) in comparison to the ProbFlowFields method on the Middlebury benchmark. †Please note that we did not re-evaluate FlowFields, but show the publicly available results.

| Uncertainty measure | AUC | rel. chg. | CC | rel. chg. |
|---|---|---|---|---|
| Gradient [3] | 1.244 | 1.72 | -0.077 | 1.21 |
| Laplace | 0.539 | 0.18 | 0.297 | 0.20 |
| Energy [8] | 0.563 | 0.23 | 0.253 | 0.32 |
| Learned [30] | 0.473 | 0.04 | **0.374** | **>-0.01** |
| ProbFlowFields *(ours)* | **0.457** | **0.00** | 0.371 | 0.00 |
| Oracle | *0.247* | – | *1.000* | – |

Table 10. Area under curve (AUC), Spearman's rank correlation coefficient (CC), and relative change (rel. chg.) in comparison to the energy uncertainty measure on the Middlebury training set.

| Average endpoint error | avg. rank | Army (Hidden texture) GT im0 im1 | | | Mequon (Hidden texture) GT im0 im1 | | | Schefflera (Hidden texture) GT im0 im1 | | | Wooden (Hidden texture) GT im0 im1 | | | Grove (Synthetic) GT im0 im1 | | | Urban (Synthetic) GT im0 im1 | | | Yosemite (Synthetic) GT im0 im1 | | | Teddy (Stereo) GT im0 im1 | | |
|---|---|---|---|---|---|---|---|---|---|---|---|---|---|---|---|---|---|---|---|---|---|---|---|---|---|
| | | all | disc | untext | all | disc | untext | all | disc | untext | all | disc | untext | all | disc | untext | all | disc | untext | all | disc | untext | all | disc | untext |
| NNF-Local [87] | 3.4 | **0.07** 1 | 0.20 2 | **0.05** 1 | 0.15 2 | 0.51 4 | 0.12 5 | 0.18 2 | 0.37 2 | 0.14 2 | 0.10 2 | 0.49 5 | 0.06 2 | **0.41** 1 | **0.61** 1 | 0.21 2 | 0.23 2 | 0.66 3 | 0.19 2 | 0.10 6 | 0.12 12 | 0.17 13 | 0.34 2 | 0.80 6 | 0.23 2 |
| PMMST [114] | 9.3 | 0.09 32 | 0.21 5 | 0.07 16 | 0.18 11 | 0.51 4 | 0.16 28 | 0.21 9 | 0.42 8 | 0.17 16 | 0.10 2 | **0.33** 1 | 0.08 13 | 0.51 5 | 0.74 4 | 0.28 7 | 0.24 3 | 0.65 2 | 0.20 5 | 0.11 18 | 0.12 12 | 0.17 13 | 0.37 4 | 0.74 2 | 0.35 4 |
| OFLAF [77] | 9.8 | 0.08 9 | 0.21 5 | 0.06 6 | 0.16 6 | 0.53 6 | 0.12 5 | 0.19 3 | 0.37 2 | 0.14 2 | 0.14 9 | 0.77 29 | 0.07 5 | 0.51 5 | 0.78 7 | 0.25 4 | 0.31 11 | 0.76 4 | 0.25 16 | 0.11 18 | 0.12 12 | 0.21 40 | 0.42 10 | 0.78 4 | 0.63 18 |
| MDP-Flow2 [68] | 10.6 | 0.08 9 | 0.21 5 | 0.07 16 | 0.15 2 | **0.48** 1 | **0.11** 1 | 0.20 5 | 0.40 5 | 0.14 2 | 0.15 22 | 0.80 37 | 0.08 13 | 0.63 19 | 0.93 19 | 0.43 20 | 0.26 5 | 0.76 4 | 0.23 9 | 0.11 18 | 0.12 12 | 0.17 13 | 0.38 6 | 0.79 5 | 0.44 6 |
| NN-field [71] | 11.8 | 0.08 9 | 0.22 17 | **0.05** 1 | 0.17 8 | 0.55 10 | 0.13 11 | 0.19 3 | 0.39 4 | 0.15 7 | **0.09** 1 | 0.48 4 | **0.05** 1 | **0.41** 1 | **0.61** 1 | **0.20** 1 | 0.52 60 | **0.64** 1 | 0.26 19 | 0.13 42 | 0.13 36 | 0.20 34 | 0.35 3 | 0.83 8 | **0.21** 1 |
| ComponentFusion [96] | 13.9 | **0.07** 1 | 0.21 5 | **0.05** 1 | 0.16 6 | 0.55 10 | 0.12 5 | 0.20 5 | 0.44 9 | 0.15 7 | 0.11 4 | 0.65 9 | 0.06 2 | 0.71 35 | 1.07 40 | 0.53 37 | 0.32 15 | 1.06 27 | 0.28 22 | 0.11 18 | 0.13 36 | 0.15 8 | 0.41 9 | 0.88 13 | 0.54 9 |
| TC/T-Flow [76] | 19.5 | **0.07** 1 | 0.21 5 | **0.05** 1 | 0.19 18 | 0.68 34 | 0.12 5 | 0.28 26 | 0.66 31 | 0.14 2 | 0.14 9 | 0.86 47 | 0.07 5 | 0.67 29 | 0.98 29 | 0.49 31 | **0.22** 1 | 0.82 9 | 0.19 2 | 0.11 18 | 0.11 2 | 0.30 85 | 0.50 27 | 1.02 30 | 0.64 20 |
| WLIF-Flow [93] | 19.8 | 0.08 9 | 0.21 5 | 0.06 6 | 0.18 11 | 0.55 10 | 0.15 23 | 0.25 18 | 0.56 20 | 0.17 16 | 0.14 9 | 0.68 10 | 0.08 13 | 0.61 16 | 0.91 17 | 0.41 18 | 0.43 35 | 0.96 15 | 0.29 28 | 0.13 42 | 0.12 12 | 0.21 40 | 0.51 32 | 1.03 33 | 0.72 38 |
| NNF-EAC [103] | 21.3 | 0.09 32 | 0.22 17 | 0.07 16 | 0.17 8 | 0.53 6 | 0.13 11 | 0.23 11 | 0.49 12 | 0.15 7 | 0.16 36 | 0.80 37 | 0.09 28 | 0.60 13 | 0.89 13 | 0.40 16 | 0.38 25 | 0.78 6 | 0.28 22 | 0.12 31 | 0.12 12 | 0.18 26 | 0.57 45 | 1.24 49 | 0.69 32 |
| Layers++ [37] | 21.9 | 0.08 9 | 0.21 5 | 0.07 16 | 0.19 18 | 0.56 13 | 0.17 35 | 0.20 5 | 0.40 5 | 0.18 27 | 0.13 8 | 0.58 7 | 0.07 5 | 0.48 3 | 0.70 3 | 0.33 9 | 0.47 47 | 1.01 19 | 0.33 47 | 0.15 65 | 0.14 58 | 0.24 53 | 0.46 17 | 0.88 13 | 0.72 38 |
| LME [70] | 22.9 | 0.08 9 | 0.22 17 | 0.06 6 | 0.15 2 | 0.49 2 | **0.11** 1 | 0.30 35 | 0.64 26 | 0.31 89 | 0.15 22 | 0.78 33 | 0.09 28 | 0.66 25 | 0.96 24 | 0.53 37 | 0.33 16 | 1.18 44 | 0.28 22 | 0.12 31 | 0.12 12 | 0.18 26 | 0.44 12 | 0.91 16 | 0.61 14 |
| IROF++ [58] | 23.0 | 0.08 9 | 0.23 24 | 0.07 16 | 0.21 32 | 0.68 34 | 0.17 35 | 0.28 26 | 0.63 24 | 0.19 39 | 0.15 22 | 0.73 22 | 0.09 28 | 0.60 13 | 0.89 13 | 0.42 19 | 0.43 35 | 1.08 30 | 0.31 37 | 0.10 6 | 0.12 12 | 0.12 4 | 0.47 19 | 0.98 23 | 0.68 31 |
| nLayers [57] | 23.8 | **0.07** 1 | **0.19** 1 | 0.06 6 | 0.22 40 | 0.59 16 | 0.19 57 | 0.25 18 | 0.54 16 | 0.20 48 | 0.15 22 | 0.84 43 | 0.08 13 | 0.53 7 | 0.78 7 | 0.34 11 | 0.44 39 | 0.84 10 | 0.30 33 | 0.13 42 | 0.13 36 | 0.20 34 | 0.47 19 | 0.97 22 | 0.67 29 |
| HAST [109] | 25.1 | **0.07** 1 | 0.20 2 | **0.05** 1 | 0.18 11 | 0.54 8 | 0.13 11 | **0.17** 1 | **0.32** 1 | **0.12** 1 | 0.15 22 | 0.90 58 | 0.06 2 | 0.49 4 | 0.74 4 | 0.22 3 | 0.58 70 | 1.09 31 | 0.44 70 | 0.19 96 | 0.17 87 | 0.47 114 | **0.32** 1 | **0.64** 1 | 0.33 3 |
| PH-Flow [101] | 25.8 | 0.08 9 | 0.24 31 | 0.07 16 | 0.21 32 | 0.68 34 | 0.17 35 | 0.23 11 | 0.49 12 | 0.19 39 | 0.16 36 | 0.83 41 | 0.09 28 | 0.56 9 | 0.83 9 | 0.38 13 | 0.30 9 | 0.81 7 | 0.24 14 | 0.15 65 | 0.13 36 | 0.30 85 | 0.43 11 | 0.85 9 | 0.66 27 |
| FC-2Layers-FF [74] | 25.8 | 0.08 9 | 0.21 5 | 0.07 16 | 0.21 32 | 0.70 42 | 0.17 35 | 0.20 5 | 0.40 5 | 0.18 27 | 0.15 22 | 0.76 28 | 0.08 13 | 0.53 7 | 0.77 6 | 0.37 12 | 0.49 53 | 1.02 20 | 0.33 47 | 0.16 76 | 0.13 36 | 0.29 80 | 0.44 12 | 0.87 12 | 0.64 20 |
| Correlation Flow [75] | 26.5 | 0.09 32 | 0.23 24 | 0.07 16 | 0.17 8 | 0.58 15 | **0.11** 1 | 0.43 65 | 0.99 67 | 0.15 7 | 0.11 4 | 0.47 3 | 0.08 13 | 0.75 41 | 1.08 41 | 0.56 42 | 0.41 31 | 0.92 13 | 0.30 33 | 0.14 52 | 0.13 36 | 0.27 69 | 0.40 8 | 0.85 9 | 0.42 5 |
| AGIF+OF [85] | 28.0 | 0.08 9 | 0.22 17 | 0.07 16 | 0.23 55 | 0.73 47 | 0.18 47 | 0.28 26 | 0.66 31 | 0.18 27 | 0.14 9 | 0.70 13 | 0.08 13 | 0.57 10 | 0.85 10 | 0.38 13 | 0.47 47 | 0.97 16 | 0.31 37 | 0.13 42 | 0.13 36 | 0.22 46 | 0.51 32 | 0.99 26 | 0.74 47 |
| RNLOD-Flow [121] | 28.0 | **0.07** 1 | 0.20 2 | 0.06 6 | 0.19 18 | 0.68 34 | 0.13 11 | 0.33 49 | 0.79 50 | 0.17 16 | 0.14 9 | 0.73 22 | 0.07 5 | 0.69 33 | 1.03 33 | 0.48 27 | 0.37 24 | 0.99 17 | 0.29 28 | 0.16 76 | 0.16 79 | 0.29 80 | 0.45 14 | 0.88 13 | 0.65 25 |
| ProbFlowFields [128] | 28.8 | 0.10 47 | 0.31 73 | 0.08 44 | 0.19 18 | 0.63 23 | 0.17 35 | 0.27 22 | 0.63 24 | 0.22 59 | 0.11 4 | 0.49 5 | 0.07 5 | 0.82 51 | 1.22 56 | 0.59 45 | 0.25 4 | 1.05 26 | 0.21 6 | 0.09 4 | 0.12 12 | 0.17 13 | 0.58 47 | 1.33 51 | 0.62 16 |
| ⋮ | | | | | | | | | | | ⋮ | | | | | | | | | | | | ⋮ | | |
| DeepFlow [86] | 65.9 | 0.12 79 | 0.31 73 | 0.11 91 | 0.28 81 | 0.82 72 | 0.22 82 | 0.44 71 | 1.00 68 | 0.33 90 | 0.26 86 | 1.34 96 | 0.15 81 | 0.81 49 | 1.21 53 | 0.58 44 | 0.38 25 | 1.55 78 | 0.25 16 | 0.11 18 | 0.11 2 | 0.24 53 | 0.93 91 | 1.82 93 | 1.12 89 |
| ProbClassicA [127] | 66.5 | 0.10 47 | 0.28 54 | 0.08 44 | 0.22 40 | 0.78 61 | 0.16 28 | 0.57 88 | 1.17 84 | 0.22 59 | 0.21 72 | 1.24 86 | 0.11 56 | 0.86 58 | 1.30 72 | 0.63 52 | 0.55 65 | 1.74 93 | 0.37 66 | 0.16 76 | 0.14 58 | 0.34 99 | 0.81 82 | 1.74 85 | 0.93 71 |
| TriangleFlow [30] | 66.7 | 0.11 59 | 0.29 62 | 0.09 65 | 0.26 73 | 0.95 88 | 0.17 35 | 0.47 74 | 1.07 72 | 0.18 27 | 0.16 36 | 0.87 51 | 0.09 28 | 1.07 97 | 1.47 104 | 1.10 98 | 0.87 86 | 1.39 65 | 0.57 87 | 0.15 65 | 0.19 109 | 0.23 52 | 0.63 55 | 1.33 51 | 0.84 61 |

Figure 6. Screenshot of private Middlebury table showing results for ProbFlowFields and ProbClassicA (status as of July 2017).

dataset for our ProbFlowFields approach. The results evaluating the AEPE on the Middlebury benchmark can be found in Table 9. We outperform the original FlowFields approach on training and test and obtain improved results in comparison to FlowFields$^{\star}$. Please note that the Middlebury benchmark policy allows no more than one entry per method in the public table. Therefore, we decided to show the results of ProbFlowFields on the Middlebury website whereas the results of ProbClassicA from Table 1 of the main paper are only visible in a private table, see Fig. 6 for a screenshot.

Table 10 shows an evaluation of different uncertainty measures. In contrast to our remaining experiments, the Laplace and learned uncertainty measures both outperform the energy-based approach. Our uncertainty measure is slightly outperformed by $\Psi_{\text{Learned}}$ w.r.t. the CC metric. However, ProbFlowFields shows clearly superior results considering the AUC.